\documentclass[letterpaper]{article} % DO NOT CHANGE THIS
\usepackage{aaai2026}  % DO NOT CHANGE THIS
\usepackage{times}  % DO NOT CHANGE THIS
\usepackage{helvet}  % DO NOT CHANGE THIS
\usepackage{courier}  % DO NOT CHANGE THIS
\usepackage[hyphens]{url}  % DO NOT CHANGE THIS
\usepackage{graphicx} % DO NOT CHANGE THIS
\usepackage{natbib}  % DO NOT CHANGE THIS AND DO NOT ADD ANY OPTIONS TO IT
\usepackage{caption} % DO NOT CHANGE THIS AND DO NOT ADD ANY OPTIONS TO IT
\usepackage{algorithm}
\usepackage{algorithmic}
\usepackage{makecell}
\usepackage{booktabs}
\usepackage{tabularx}
\usepackage{relsize}
\usepackage{multirow}

\usepackage{amsmath, bm}
\usepackage{amssymb}

\usepackage{graphicx} 
\usepackage{subfig}
\usepackage{xspace}
\usepackage{soul}

\def\onedot{.\xspace}
\def\eg{\emph{e.g}\onedot} 
\def\ie{\emph{i.e}\onedot}

\def\etal{\emph{et al}\onedot}
\usepackage{newfloat}
\usepackage{listings}
\DeclareCaptionStyle{ruled}{labelfont=normalfont,labelsep=colon,strut=off} % DO NOT CHANGE THIS
\floatstyle{ruled}
\newfloat{listing}{tb}{lst}{}
\floatname{listing}{Listing}
\nocopyright
\title{ElastoGen: 4D Generative Elastodynamics}
\author{
    Yutao Feng\textsuperscript{\rm 1,\rm 2}\equalcontrib,\quad Yintong Shang\textsuperscript{\rm 1}\equalcontrib,\quad Xiang Feng\textsuperscript{\rm 1,\rm 2}\equalcontrib,\quad Lei Lan\textsuperscript{\rm 1},\quad Shandian Zhe\textsuperscript{\rm 1},\quad Tianjia Shao\textsuperscript{\rm 2}\thanks{Corresponding author},\quad Hongzhi Wu\textsuperscript{\rm 2},\quad Kun Zhou\textsuperscript{\rm 2},\quad Chenfanfu Jiang\textsuperscript{\rm 3},\quad Yin Yang\textsuperscript{\rm 1}
}
\affiliations{
    \textsuperscript{\rm 1}University of Utah\quad\quad
    \textsuperscript{\rm 2}State Key Laboratory of CAD\&CG, Zhejiang University\quad\quad
    \textsuperscript{\rm 3}UCLA
    
}

\usepackage{bibentry}
\begin{document}

\maketitle

\begin{abstract}
We present ElastoGen, a \emph{knowledge-driven} AI model that generates physically accurate 4D elastodynamics. Unlike deep models that learn from video- or image-based observations,  
ElastoGen leverages the principles of physics and learns from established mathematical and optimization procedures. The core idea of ElastoGen is converting the differential equation, corresponding to the nonlinear force equilibrium, into a series of iterative local convolution-like operations, which naturally fit deep architectures. We carefully build our network module following this overarching design philosophy. ElastoGen is much more lightweight in terms of both training requirements and network scale than deep generative models. Because of its alignment with actual physical procedures, ElastoGen efficiently generates accurate dynamics for a wide range of hyperelastic materials and can be easily integrated with upstream and downstream deep modules to enable end-to-end 4D generation.
\end{abstract}
\section{Introduction}
Recent advancements in generative models have enhanced the ability to produce high-quality digital contents across diverse media formats (e.g. images, videos, 3D models, 4D data). In particular, the generation of 4D data, including both spatial and temporal dimensions, has seen notable progress~\citep{singer2023text,shen2023make,xu2024comp4d,ling2023align,bahmani2024tc4d,yin20234dgen,bah20244dfy}. 

On the other hand, learning physical dynamics that exhibit temporal consistency and adhere to physical laws from observable data remains a difficult problem. Data are in the wild and noisy. Their underlying coherence is agnostic to the user. As a result, existing deep models have to assume some distributions of the data, which may not be the case in reality. In theory, the network would extract any knowledge provided sufficient data. In practice however, such data-based learning becomes more and more cumbersome with increased dimensionality of generated contents -- it is unintuitive to define the right network structure to guide a physically meaningful generation; it requires terabyte- or petabyte-scale high-quality training data, and center-level computing resource to facilitate the training. Those theoretical and practical obstacles combined impose significant challenges.

We explore a new way to establish physics-in-the-loop generative models. Our argument is that \textit{learning from knowledge} instead of from raw data is more effective for generative models. Physical laws and principles are often in the form of partial differential equations (PDEs) and numerically solved with discretized differential operators. We note that those operators hold a similar structure as a convolution kernel on the problem domain, where the values of those convolution kernels depend on the specific problem setting. Inspired by those observations, we propose ElastoGen, a knowledge-driven neural model that generates physically accurate and coherent 4D elastodynamics. ElastoGen can be easily coupled and integrated with upstream and downstream neural modules to enable end-to-end 4D generation. The core idea of ElastoGen is converting the global differential operator, corresponding to the nonlinear force equilibrium, to iterative local convolution-like procedures. Such knowledge-level priors allow us to design dedicated network modules for ElastoGen, where each network module has a well-defined purpose of relaxing locally concentrated strain rather than being treated as a piece of a black box. Compared with other data-learning-based generative models, ElastoGen is lightweight -- in terms of both training requirements and the network scales. Furthermore, due to its consistency with physics procedure, ElastoGen generates physically accurate dynamics for a wide range of hyperelastic materials. Specifically, we summarize some features of ElastoGen as follows:

\noindent \textbf{Compact generative network inspired by physics principles}\hspace{5 pt}
The network architecture of ElastoGen is strongly inspired by our prior knowledge of physics and corresponding numerical procedures. This allows a compact and effective generative framework in the form of deep neural networks. The training efforts for such a carefully tailored deep model become lightweight as well.

\noindent \textbf{NeuralMTL with diffusion parameterization}\hspace{5 pt}
ElastoGen features a neural material module, \emph{NeuralMTL}, to encode the underlying constitutive relations for real-world hyperelastic materials such as Neo-Hookean and or Saint Venant-Kirchhoff (StVK). We leverage a lightweight conditional diffusion model to predict its network parameters to isolate our training efforts.

\noindent \textbf{Nested RNN with low-frequency encoding}\hspace{5pt}
ElastoGen constitutes a two-level recurrent neural network (RNN) architecture. An encoder extracts low-frequency deformations so that the inner RNN relaxation only takes care of the local high-frequency strains. This design makes ElastoGen more efficient for stiff materials.  

\section{Related work}

\noindent \textbf{4D generation with diffusion models}
The primary objective of generative models is to produce new, high-quality samples from vast datasets. These models are designed to learn and understand the distribution of data, thereby generating samples that meet specific criteria. Recently, diffusion models have emerged as a powerful technique, achieving state-of-the-art results in generating high-fidelity images~\citep{Sohl-DicksteinW2015diffusion, HoJA2020ddpm, rombach2022high}, beating the competitors that are based on generative adversarial networks (GANs)~\citep{goodfellow2014gan,zhu2019dm,chan2021pi} or variational autoencoders (VAEs)~\citep{kingma2014vae,Child2021vae,razavi2019generating}. This sets the stage for further explorations in more complex applications, such as the generation of 3D content~\citep{jain2022zero, lin2023magic3d, metzer2023latent, poole2022dreamfusion, wang2024prolificdreamer, liu2023zero, liu2023one2345, feng2024arm}, video content~\citep{blattmann2023align, harvey2022flexible, ho2022video, ho2022imagen, karras2023dreampose, ni2023conditional}, and 3D videos or called 4D dynamic scenes (3D objects with shape change throughout time)~\citep{singer2023text,shen2023make,xu2024comp4d,ling2023align,bahmani2024tc4d,yin20234dgen,bah20244dfy}. These advanced applications demonstrate the versatility and expanding potential of diffusion models across diverse media formats. However, existing 4D generation techniques struggle to ensure temporal consistency and require substantial training data, underscoring the challenges of capturing and replicating the dynamic and interconnected behaviors present in real-world scenarios within a generative model framework.

\noindent \textbf{Neural physical dynamics}
Physical dynamics has traditional numerical solutions, such as the finite element method (FEM)~\citep{zienkiewicz1971finite,zienkiewicz2005finite} and mass-spring systems~\citep{liu2013pd}. Each approach offers distinct advantages and limitations. For example, Position-Based Dynamics (PBD)~\citep{muller2007pbd} and Projective Dynamics (PD)~\citep{Bouaziz2014PD} offer simplified implementation and faster convergence but can struggle with complex material behaviors and do not always guarantee consistent convergence rates. Recently, neural physics solvers, which integrate neural networks with traditional solvers, aim to accelerate and simplify the computation process. The pioneering works~\citep{chang2017object, battaglia2016interaction} directly utilized neural networks to predict dynamics, achieving promising results in simple particle systems. Subsequent studies~\citep{sanchez2018graph, kipf2018neural, ajay2018augmenting, li2019propagation, li2019learning, li2019particle} adopted network architectures to the specific features of the systems, thereby enhancing performance. Differentiable simulators are also proposed~\cite{degrave2019differentiable,de2018end,hu2019difftaichi}, enabling end-to-end training of the physics-based behavior from ground-truth data.
The advent of Physics Informed Neural Networks (PINNs)~\citep{raissi2019physics, pakravan2021solving} marks a leap forward. These networks incorporate extensive physical information to constrain and guide the learning process, ensuring that predictions adhere more closely to physical laws and has succeeded in domains such as cloths~\citep{Geng2020physical} and fluids~\citep{um2020solver, gibou2019sharp, chu2022fpinn}. Some work~\citep{yang2020learning} shifts away from end-to-end structures and use neural networks to optimize part of the simulation. 
Another line of research generates dynamics through physics-based simulators, where network learns static information while physical laws govern the generation of dynamics~\citep{li2023pac, feng2023pie, xie2023physgaussian, feng2024gaussian, jiang2024vr}, giving physical meanings to Neural Radiance Fields (NeRF)~\citep{mildenhall2020nerf, kerbl20233Dgaussians}. These methods demonstrate the benefits of embedding human knowledge into networks to reduce the learning burden.

\section{Background}\label{sec:bg}
To make the paper self-contained, we start with a brief review of the preliminaries of elastodynamics and the diffusion model.

\subsection{Variational optimization of elastodynamics}
The dynamic equilibrium of a 3D model can be characterized by $\frac{\mathrm{d} }{\mathrm{d} t}
\left(\frac{\partial L}{\partial \dot{\mathbf{q}}}\right) - \frac{\partial L}{\partial \mathbf{q}} = \mathbf{f}_q$,
% \begin{equation}\label{eq:lagrangian}
%\frac{\mathrm{d} }{\mathrm{d} t}
%\left(\frac{\partial L}{\partial \dot{\mathbf{q}}}\right) - \frac{\partial L}{\partial \mathbf{q}} = \mathbf{f}_q,
% \end{equation}
where $L = T - U$ is system \emph{Lagrangian} \ie, the difference between the kinematic energy ($T$) and the potential energy ($U$). $\mathbf{q}$ and $\dot{\mathbf{q}}$ are generalized coordinate and velocity. $\mathbf{f}_q$ is the generalized external force. 
With the implicit Euler time integration scheme: $\mathbf{q}^{n+1} = \mathbf{q}^n + h\dot{\mathbf{q}}^{n+1}$, $\dot{\mathbf{q}}^{n+1} = \dot{\mathbf{q}}^n + h\ddot{\mathbf{q}}^{n+1}$, it can be reformulated as a nonlinear optimization to be solved at each time step:
% \begin{equation}\label{eq:totEnergy}
% \mathbf{q}_{n+1} =\underset{\mathbf{q}}{\mathrm{argmin}}\left\{ \frac{1}{2h^2}\|\mathbf{q} - \mathbf{q}_n - h\dot{\mathbf{q}}_n\|^2_{\mathbf{M}}-\mathbf{q}^\top\mathbf{f}_q+U(\mathbf{q})\right\},
% \end{equation}
\begin{equation}\label{eq:totEnergy}
\mathbf{q}^{n+1} =\underset{\mathbf{q}}{\mathrm{argmin}}\left\{ \frac{1}{2h^2}\|\mathbf{q} - \mathbf{\hat{q}}\|^2_{\mathbf{M}}+U(\mathbf{q})\right\},
\end{equation}
where $\mathbf{\hat{q}} = \mathbf{q}^n - h\dot{\mathbf{q}}^n-h^2\mathbf{M}^{-1}\mathbf{f}_q$, the superscript $n$ and $n+1$ indicates the time step, $h$ is the time step size, and $\mathbf{M}$ is the mass matrix. 

\subsection{Diffusion model}
\label{sec:bg_diffusion}
A diffusion model transforms a probability from the real data distribution $ \mathcal{P}_{\text{real}} $ to a target distribution $\mathcal{P}_{\text{target}}$ through diffusion and denoising.

\noindent \textbf{Diffusion.}\hspace{5 pt}
The diffusion process incrementally adds Gaussian noise to the initial data $\mathbf{x}_0 \sim \mathcal{P}_{\text{target}}$, gradually transforming it into a sequence $\mathbf{x}_1, \mathbf{x}_2, ..., \mathbf{x}_T$, where $\mathbf{x}_T$ approximates the real distribution $\mathcal{P}_{\text{real}}$. The aim is to learn a noise prediction model $ \boldsymbol{\epsilon}_{\theta}(\mathbf{x}_t, t)$, estimating the noise at each iteration $t$ to facilitate data recovery in the denoising phase. The noise learning objective is formulated as:
\begin{equation}\label{eq:diffusion}
L = \mathbb{E}_{\mathbf{x}_0 \sim \mathcal{P}_{\text{target}}, \boldsymbol{\epsilon} \sim \mathcal{N}(\mathbf{0}, \mathbf{I}), t\sim \text{Uniform}(\{1, ..., T\})} [\|\boldsymbol{\epsilon} - \boldsymbol{\epsilon}_{\theta}(\mathbf{x}_t, t)\|^2].
\end{equation}
%where $\|\cdot\|_2^2$ denotes the mean squared error loss.

\noindent \textbf{Denoising.}\hspace{5 pt}
Denoising iteratively removes noise from $\mathbf{x}_T \sim \mathcal{P}_{\text{real}}$, recovering the original data $\mathbf{x}_0$ by adjusting the noisy data at each iteration $t$ as:
\begin{equation}\label{eq:denoising}
\mathbf{x}_{t-1} = \frac{1}{\sqrt{\alpha_t}}(\mathbf{x}_t - \frac{1-\alpha_t}{\sqrt{(1 - \overline{\alpha}_t)}}\boldsymbol{\epsilon}_{\theta}(\mathbf{x}_t, t)) + \sigma_t \mathbf{z},  \mathbf{z} \sim N(\mathbf{0}, \mathbf{I}),
\end{equation}
where $1-\alpha_t =\beta_t$ is a scheduled variance, and $\sigma_t$ is typically set to $\sigma_t = \sqrt{\beta_t}$. $N(\mathbf{0}, \mathbf{I})$ is standard normal distribution.
Diffusion and denoising processes allow for effective modeling of the transition between distributions, using learned Gaussian transitions for noise prediction and reduction.
\section{Methodology}\label{sec:method}
\begin{figure*}[ht]
\centering
\begin{tabular}{c@{}c}
\includegraphics[width=0.5\linewidth]{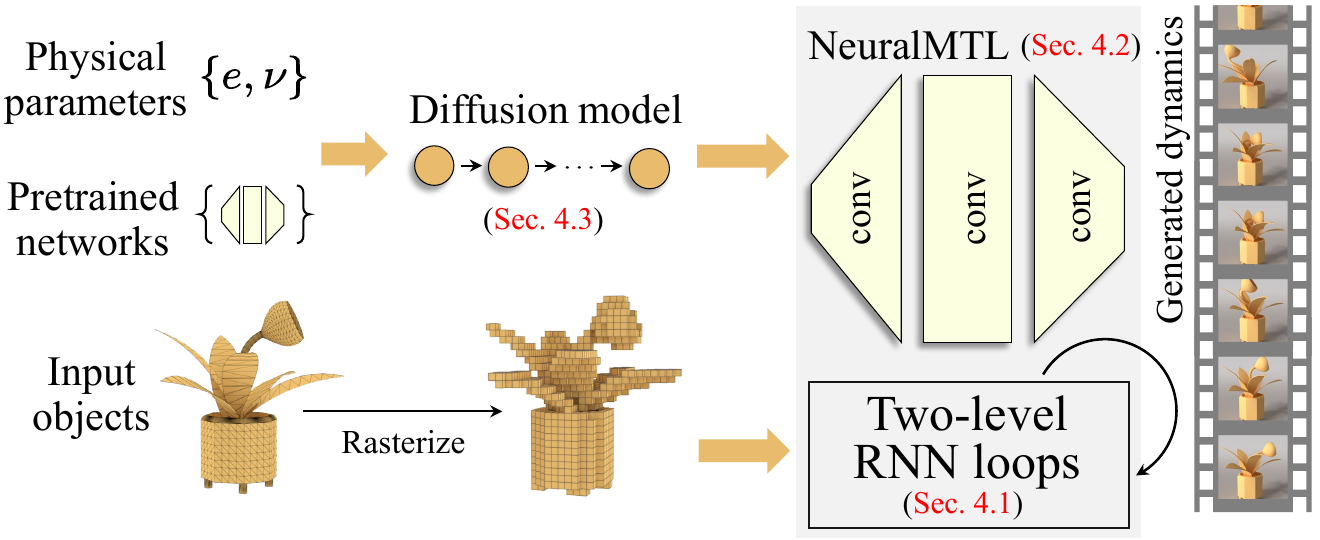} & 
\includegraphics[width=0.5\linewidth]{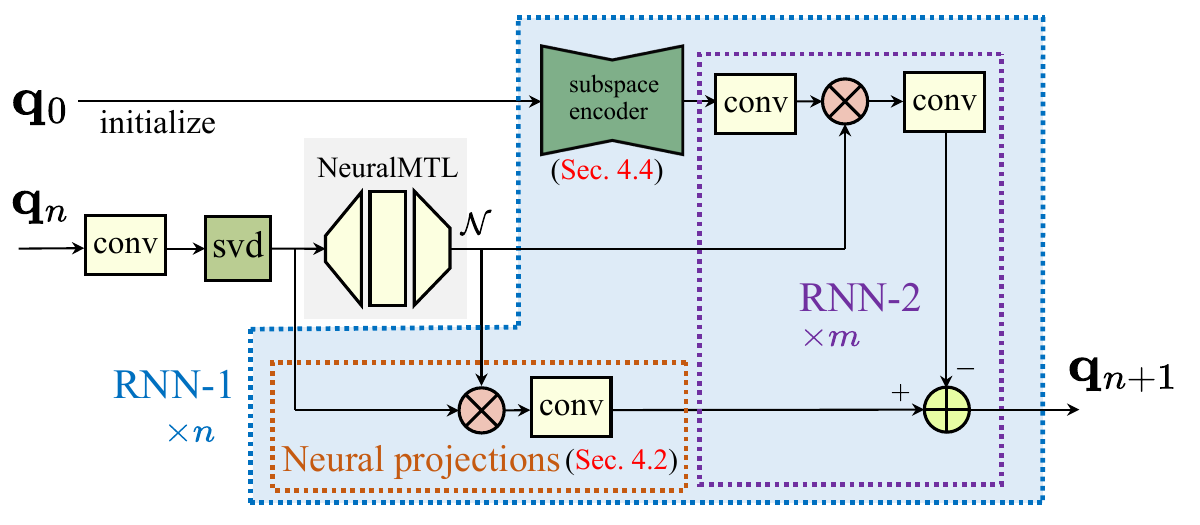} \\
\small{\textbf{(a)} The pipeline of ElastoGen.} &
\small{\textbf{(b)} The network structure.} \\
\end{tabular}
% \vspace{-0.2cm}
 \caption{\textbf{Pipeline overview.}~\textbf{(a)} ElastoGen rasterizes an input 3D model (with boundary conditions) and generates parameters filling our NeuralMTL module. Conceptually, NeuralMTL predicts locally concentrated strain of the object, which is relaxed by a nested RNN loop. \textbf{(b)} The RNN predicts the future trajectory of the object. There are two sub RNN modules. RNN-1 repeatedly relaxes the local stress in a 3D convolution manner. Those relaxed strains are converted to positional signals, and RNN-2 merges local deformation into a displacement field of the object. ElastoGen automatically checks the accuracy of the prediction of both RNN loops, and outputs the final prediction of \(\mathbf{q}_{n+1}\) once the prediction error reaches the prescribed threshold.}
\label{fig:pipeline}
\end{figure*}

ElastoGen, as illustrated in Fig.\ref{fig:pipeline}, is a lightweight generative deep model that produces physically grounded 4D content from general object descriptions, such as stiffness or mass. ElastoGen rasterizes the input shape and uses a nested two-level recurrent neural network (RNN) to predict its trajectory sequentially. Each prediction undergoes an accuracy check to ensure physical validity. The network design follows a numerical procedure based on variational optimization as in Eq.\eqref{eq:totEnergy}, ensuring that ElastoGen avoids redundant components that could lead to overfitting. The following sections detail each major module of the pipeline.

\subsection{Method overview: piece-wise local quadratic approximation}

Our elastodynamics generation mimics numerical optimization procedures that minimize the variational energy as in Eq.~\eqref{eq:totEnergy}. It is possible to tackle this problem at the global level, \ie, optimizing all the degrees of freedom (DoFs) of the system at once e.g., using Newton's method. Such a brute-force scheme requires to learn dense inter-correlations among features at all DoFs, which inevitably leads to complex and large-scale network architectures with numerous parameters to be learned. 

Alternatively, we opt for a divide-and-conquer way to approach Eq.~\eqref{eq:totEnergy}. We consider the total potential energy $U$ as the summation of multiple energies of quadratic form: $U(\mathbf{q})\approx\sum_i E_i(\mathbf{q}_i)$,  where $E_i(\mathbf{q}_i)=\min_{\mathbf{p}_i\in \mathcal{M}_i} \frac{\omega_i}{2}\left\|\mathcal{G}_i [\mathbf{q}_i] -\mathbf{p}_i\right\|^2$. Here, $i$ indicates the $i$-th sub-volume of the object. For instance, one may discretize the object into a tetrahedral mesh, and $E_i$ then represents the elastic potential stored at the $i$-th element. $\mathcal{G}_i$ denotes a \emph{discrete differential operator}, which converts positional features $\mathbf{q}_i$ to strain-level features. To this end, we build $\mathcal{G}_i$ such that $\mathcal{G}_i[\mathbf{q}_i] = \text{vec}(\mathbf{F}_i)$, i.e., the vectorized deformation gradient ($\mathbf{F}\in\mathbb{R}^{3 \times 3}$) of the sub-volume, which gives the local first-order approximation of the displacement field. The constraint manifold $\mathcal{M}_i$ denotes the zero level set of $E_i$. In other words, we consider $E_i$ as a quadratic energy based on how far local displacement $\mathbf{q}_i$ is from its closest energy-free configuration ($\mathbf{p}_i$), given the local material stiffness $\omega_i$. 

Provided the current deformed shape $\mathbf{q}_i$, we can find $\arg\min_{\mathbf{p}_i} \frac{\omega_i}{2}\left\|\mathcal{G}_i [\mathbf{q}_i] -\mathbf{p}_i\right\|^2$, which suggests a locally optimal descent direction to reduce $U$. The global displacement can then be obtained by minimizing $\mathbf{q}$ over $E_i$ at all the sub-volumes. 
While this is a global operation, it acts as a Laplacian-like smoothing operator, which can be approximated through repeated local smoothing. This procedure resembles shape matching methods~\citep{muller2005meshless} and PD~\citep{Bouaziz2014PD}, offering a piecewise sequential quadratic programming (SQP) approach~\citep{boggs1995sequential} to approximate $U$ locally. ElastoGen functions as a neural version of the aforementioned procedure, with a nested RNN structure that handles local strain relaxation through volume convolutions, ensuring the network remains compact and lightweight.

Unfortunately, real-world materials are more than a collection of quadratic forms. 
The elastic energy of nonlinear materials cannot be well approximated by a quadratic form of the deformation gradient. This limitation means that shape matching or PD can only handle simplified material behavior. To this end, we augment ElastoGen with a NeuralMTL module to ensure that each local SQP closely matches actual materials.

\subsection{NeuralMTL \& neural projection}
\begin{figure}[ht]
  \centering
  \begin{tabular}{c@{}c@{}c}
    \includegraphics[width=0.25\linewidth]{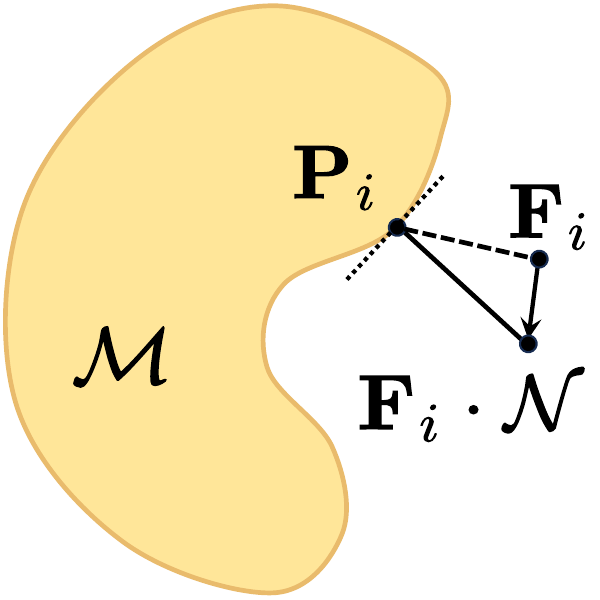} & 
    \includegraphics[width=0.25\linewidth]{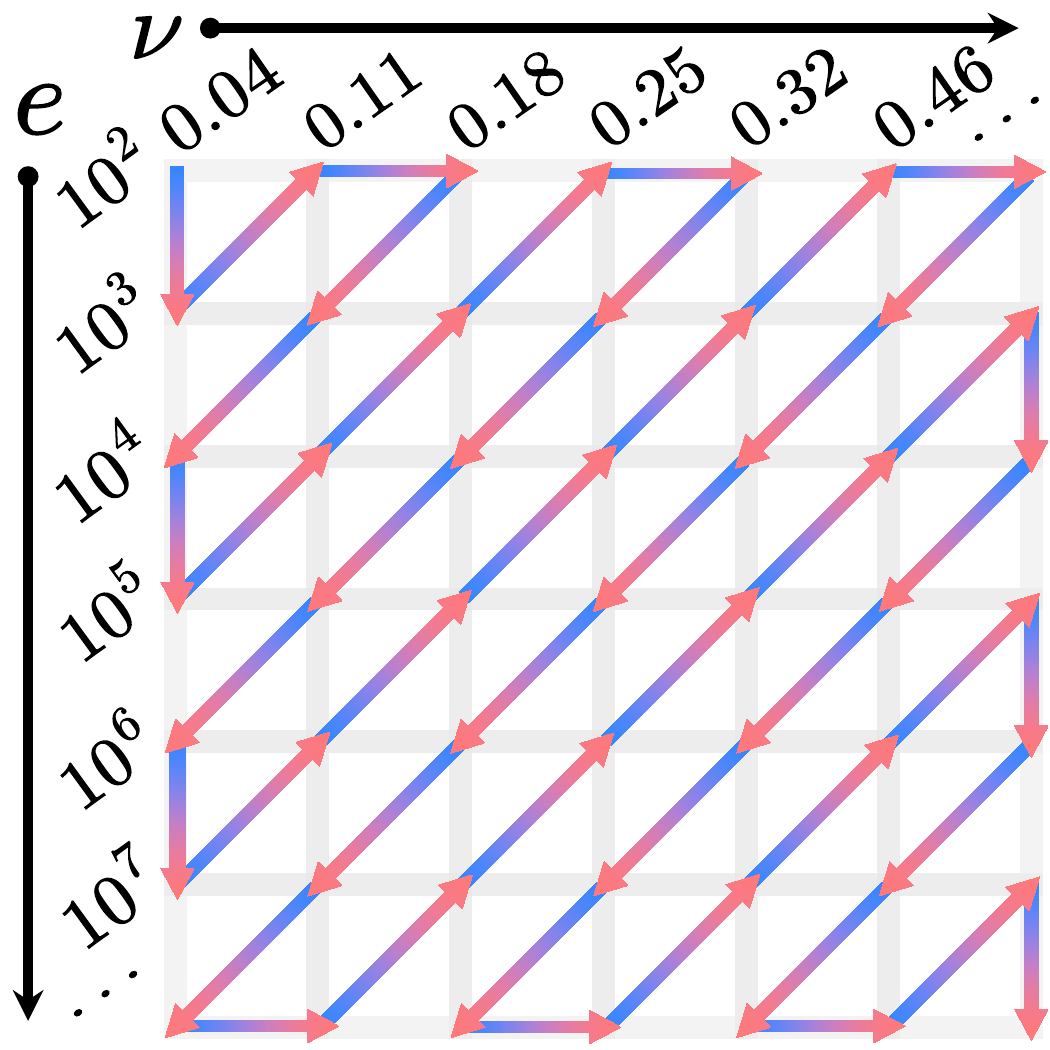} &
    \includegraphics[width=0.45\linewidth]{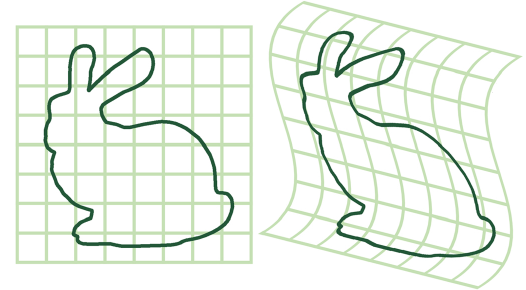}
    \\
    \small{\textbf{(a)} } &
    \small{\textbf{(b)} } & 
    \small{\textbf{(c)} }\\
    \end{tabular}
  \caption{\textbf{(a).}~NeuralMTL learns a mapping $\mathcal{N}$ to warp $\mathbf{F}_i$, enabling the quadratic strain energy to work for hyperelastic materials \textbf{(b).}~Sampling topology over 2D input space to ensure smooth output variation. \textbf{(c)}~Deforming object with rasterization grid.}
  \label{fig:illustration}
\end{figure}

The goal of NeuralMTL is to correct local quadratic approximations of $U$ so that ElastoGen faithfully generates physically accurate results for any real-world hyperelastic material. Specifically with NeuralMTL ($\mathcal{N}$), $E_i$ becomes:
\begin{equation}\label{eq:neuralmetric}
E_i(\mathbf{q}_i) = \underset{\mathbf{P}_i\in \mathcal{SO}(3)}{\mathrm{argmin}}\frac{\omega_i}{2} \left\|\mathbf{F}_i\cdot\mathcal{N}\big(\mathcal{G}_i[\mathbf{q}_i]\big) - \mathbf{P}_i\right\|^2_{F}.
\end{equation}
We set $\omega_i$ as $\omega_i=V_i e$ based on real-world material parameters: Young's modulus $e$ and the size of the sub-volume $V_i$. $\mathcal{G}_i$ extracts the deformation gradient $\text{vec}(\mathbf{F}_i)$ and feeds it to NeuralMTL, $\mathcal{N}$. As the name suggests, $\mathcal{N}$ predicts a \emph{neural strain} based on the information of local deformation $\mathbf{F}_i$. Given the material model and parameters, $\mathcal{N}$ is used for all $E_i$, and we do not put a subscript on $\mathcal{N}$. $\|\cdot\|_{F}$ denotes the Frobenius norm. $\mathcal{N}$ predicts a material-space strain prediction, which is then converted to world space by $\mathbf{F}_i$. $\mathbf{P}_i\in\mathbb{R}^{3\times3}$ is a rotation matrix i.e., $\mathbf{P}_i\in\mathcal{SO}(3)$. Intuitively, as shown in Fig.~\ref{fig:illustration}~(a), NeuralMTL warps $\mathbf{F}_i$ to a different configuration of $\mathbf{F}_i \cdot \mathcal{N}(\mathbf{F}_i)$ so that the new distance to $\mathbf{P}_i$ correctly reflects the local energy landscape of $E_i$ as visualized in the inset.

For isotropic elastic materials, we add a nonlinear singular value decomposition (SVD) activation to the operator $\mathcal{G}_i$ such that $\mathbf{F}_i=\mathbf{U}_i \mathbf{S}_i \mathbf{V}_i^\top$. $\mathbf{S}_i$ is a diagonal matrix with singular values arranged in descending order, which correspond to the local principal strains. This activation converts $E_i$ to: 
\begin{equation}\label{eq:isotropy}
\begin{split}
E_i(\mathbf{q}_i)=&\frac{\omega_i}{2} \|\mathbf{U}_i\mathbf{S}_i\mathbf{V}_i^\top\cdot\mathcal{N}(\mathcal{G}_i[\mathbf{q}_i]) - \mathbf{U}_i\mathbf{V}_i^\top\|^2_{F}\\
  =&\frac{\omega_i}{2}
  \mathrm{tr}\left(\mathbf{S}_i\mathbf{S}_i\mathbf{V}_i^\top\cdot\mathcal{N}(\mathcal{G}_i[\mathbf{q}_i])\cdot\mathcal{N}^\top(\mathcal{G}_i[\mathbf{q}_i])\mathbf{V}_i\right.\\
  &\left.+\mathbf{I}-2\mathbf{V}_i\mathbf{S}_i\mathbf{V}_i^\top\cdot\mathcal{N}(\mathcal{G}_i[\mathbf{q}_i])\right).
\end{split}
\end{equation}

We further require this learning-based strain measure that 1) NeuralMTL predicts a symmetric strain; and 2) the adjusted energy remains invariant to rotation and merely depends on $\mathbf{S}_i$. Let $\mathbf{N}_i = \mathcal{N}(\mathcal{G}_i[\mathbf{q}_i]) \in\mathbb{R}^{3 \times 3}$ be the raw output of NeuralMTL. Instead of directly imposing those restrictions during the training, we append a network module to nonlinearly activate the raw output of $\mathcal{N}$ as:
\begin{equation}
   \mathcal{N}(\mathcal{G}_i[\mathbf{q}_i]) \leftarrow \mathbf{V}_i\big(\mathbf{N}_i+\mathbf{N}_i^\top\big)\mathbf{V}_i^\top, 
\end{equation}
which further simplifies $E_i$ to:
\begin{equation}\label{eq:local}
\begin{split}
&E_i=\frac{\omega_i}{2}\mathrm{tr}\left(\mathbf{Q}_i\mathbf{Q}_i^\top\right)+ \frac{3\omega_i}{2} - \omega_i\ \mathrm{tr}\left(\mathbf{Q}_i\right),\\ &\mathbf{Q}_i(\mathbf{S}_i)=\mathbf{S}_i\left(\mathbf{N}_i+\mathbf{N}_i^\top\right).
\end{split}
\end{equation}
Intuitively, this activation escalates the order of the neural strain predicted by $\mathcal{N}$, pushing it to become a nonlinear strain estimation with 
a prescribed format --- just like upgrading an infinitesimal strain to Green's strain to better measure large rotational deformation. As a result, the neural projection corresponding to our NeuralMTL can be easily obtained as $\mathbf{P}_i = \mathbf{U}_i \mathbf{V}_i^\top$, \ie, the rotational component from $\mathbf{F}_i$. This is an important property of NeuralMTL --- if we choose to employ the network to learn an adjustment of $\mathbf{P}_i$ (which is also technically feasible), the local relaxation that predicts $\mathbf{P}_i$ becomes complicated, and the generation is less robust.

Given an input 3D object, ElastoGen rasterizes it into a set of sub-volumes. For a user-specified sub-volume, which in our implementation is a voxel that intersects with the object, $\mathcal{G}_i$ operator extracts the local covariance matrix of the displacement field over this sub-volume. Let $\mathbf{A}_i = [\mathbf{q}_1, \mathbf{q}_2,...\mathbf{q}_k]\in\mathbb{R}^{3 \times k}$ and $\bar{\mathbf{A}}_i = [\bar{\mathbf{q}}_1, \bar{\mathbf{q}}_2,...\bar{\mathbf{q}}_k]$ be deformed and rest-shape position of vertices of a sub-volume with $k$ vertices ($k = 8$ for a cubic volume). $\mathcal{G}_i$ has an analytic format of:
\begin{equation}
    \mathcal{G}_i[\mathbf{q}_i] = \left[\big(\bar{\mathbf{A}} \bar{\mathbf{A}}^\top\big)^{-1}\bar{\mathbf{A}}\otimes\mathbf{I}\right]\mathbf{q}_i,
\label{eq:analytic_operator}
\end{equation}
which is a convolutional neural network (CNN) whose weights can be pre-computed given the rasterized object. The output of $\mathcal{G}_i$ is then activated via a SVD module, which outputs $\mathbf{U}_i$, $\mathbf{V}_i$, and $\mathbf{S}_i$. 

And the NeuralMTL $\mathcal{N}$ can also be implemented as a CNN whose weights are predicted by a generative diffusion model given the material type and parameters such as Young's modulus $e$ and Poisson's ratio $\nu$. 

\subsection{Decoupling NeuralMTL from deformation and material parameters}
\label{sec:diffusion}
NeuralMTL takes $\mathbf{F}_i$ as input and outputs the neural strain measure $\mathcal{N}(\mathbf{F}_i)$. This strain is then evaluated against Eq.~\eqref{eq:neuralmetric} to check if ElastoGen reasonably minimizes Eq.~\eqref{eq:totEnergy} and is ready for the next time step. NeuralMTL is designed to accommodate material nonlinearity, meaning different material parameters $\{e, \nu\}$ result in different outputs even for the same $\mathbf{F}_i$. 

A direct approach would be training NeuralMTL $\mathcal{N}(\mathbf{F}_i, e, \nu)$ on both $\mathbf{F}_i$ and $\{e, \nu\}$. However, since NeuralMTL needs to be evaluated frequently during deformation, we decouple the influences of $\mathbf{F}_i$ and $\{e, \nu\}$ to maintain a compact network. Following the approach in~\cite{zhang2024metadiff}, we generate $\mathcal{N}(\mathbf{F}_i)$ using a diffusion model $\mathcal{D}(e, \nu)$, where $\mathbf{W}=\mathcal{D}(e, \nu)$ is the parameters for the network $\mathcal{N}(\mathbf{F}_i)$.

To train the model $\mathcal{D}$, we prepare a dataset of paired $\{e, \nu\}$ and $\mathbf{W}$. To this end, we first uniformly sample both $e$ and $\nu$ at fixed intervals and then establish a topological order, as shown in Fig.~\ref{fig:illustration}~(b). A target elastic energy $\Psi(e, \nu)$ can be easily computed for each sampled $\{e, \nu\}$. $\mathbf{W}$ is then obtained via the following optimization:
\begin{equation}\label{eq:loss}
    % \qquad\qquad\qquad\qquad
    \mathbf{W}=\underset{\mathbf{W}}{\mathrm{argmin}}\left\| \log(\Psi_{\mathcal{N}} + 1) - \log(\Psi + 1)\right\|^2,
\end{equation}
where $\Psi_{\mathcal{N}}=\frac{\omega_i}{2} \|\mathbf{F}_i\cdot \mathcal{N}( \mathbf{W}, \mathbf{F}_i) - \mathbf{U}_i\mathbf{V}_i^\top\|^2$, and $\mathcal{N}({\mathbf{W}}, \mathbf{F}_i)$ suggests parameters of $\mathcal{N}$ are prescribed by $\mathbf{W}$. 
The logarithmic function $\log$ is used to penalize energy deviations under the same deformation, ensuring non-negative energy. As the energy function varies smoothly with $\{e, \nu\}$, our predefined topological ordering of $\{e, \nu\}$ samples facilitates efficient training. $\mathbf{W}$ can converge within a few hundred gradient descent iterations when initialized from the previous $\mathbf{W}$. During inference, after $\mathcal{D}$ predicts $\mathbf{W}$, we perform a few additional gradient descent iterations to fine-tune the weights, ensuring $\mathcal{N}$ accurately fits the desired elastic energy function. This two-step process enables smooth variation of the energy function with respect to $\{e, \nu\}$, ensuring efficient and precise network parameter generation.

\subsection{Subspace encoding}
If the quadratic approximation of Eq.~\eqref{eq:totEnergy} is exact, NeuralMTL, $\mathcal{N}$, is not needed. And after obtaining $\mathbf{Q}_i$ for all voxels, we set its derivative to zero leading to: 
\begin{equation}\label{eq:global}
    \left(\frac{\mathbf{M}}{h^2} + \sum_i \mathbf{L}_i\right)\mathbf{q}^{n+1}=\mathbf{f}_q + \frac{\mathbf{M}}{h^2}(\mathbf{q}^n + h\dot{\mathbf{q}}^n) + \sum_i\mathbf{b}_i,
\end{equation}

where $\mathbf{b}_i=\mathbf{L}_i\mathbf{q}_n - \frac{\partial E_i}{\partial \mathbf{q}}$.
The term $\frac{\mathbf{M}}{h^2}+\sum_i\mathbf{L}_i$ as the \emph{global matrix}, which remains constant in this case. This allows for pre-factorization, converting the global matrix into lower and upper triangular forms to facilitate efficient solving of the linear system. However, the use of NeuralMTL introduces nonlinearities in the energy landscape, making $\mathbf{L}_i(\mathbf{q})$ dependent on the current deformed pose $\mathbf{q}$. A fully implicit evaluation would require $\nabla_{\mathbf{q}} \mathbf{L}_i$ and $\nabla_{\mathbf{q}} \mathcal{N}$, which is computationally expensive and less stable due to the need for additional training constraints (e.g., penalizing $|\nabla \mathcal{N}|$ to prevent overfitting). To address this, we adopt a lagged approach, computing $\mathbf{L}_i$ by using $\mathbf{q}$ from the most recent update.

Directly inverting the global matrix is computationally expensive due to the dense correlations between vertices. As an alternative, traditional iterative methods update each node by incorporating information from its neighbors, using results from the previous iteration, and continue relaxing locally until convergence is achieved. Inspired by these methods, we decompose the global solve for $\mathbf{q}$ by perform multiple local operations at $\mathbf{q}_i$. Additionally, since the objects are rasterized into a grid, these repetitive local operators can be efficiently implemented using a recurrent CNN. This CNN takes  $\mathbf{q}_i$ as input and outputs the relaxed result after one step. This CNN collects information from the vertices to the voxels via a convolution, then relaxes it back to the vertices using a transposed convolution. Additional implementation details can be found in our supplementary material. Conceptually, this approach functions as a matrix-free method for solving the global system, ensuring that ElastoGen generates a physically accurate trajectory.

ElastoGen is therefore built as a two-level RNN network. The outer RNN (RNN-1, see Fig.~\ref{fig:pipeline}) applies local NeuralMTL adjustments over $\mathcal{G}_i$ at each voxel region and the neural projection for $\mathbf{P}_i$. The inner RNN (RNN-2) handles global solve by iterating a CNN, approximating the local relaxation effect through repeated local operations. Each local operator relaxes the concentrated strain predicted by NeuralMTL $\mathcal{N}$ and is propagated across the object. The process iterates until the difference between the results of two consecutive iterations is below a specified threshold.

\begin{figure}
  \centering 
  \includegraphics[width = \linewidth]{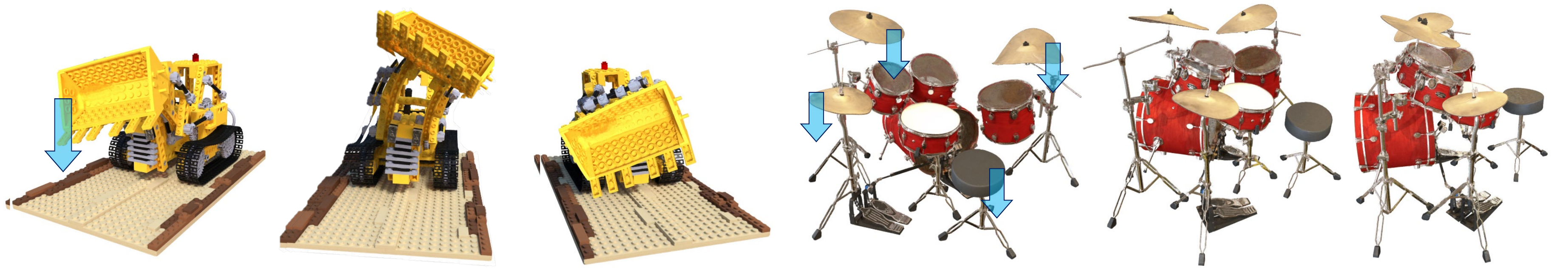}
  \caption{\textbf{ElastoGen with implicit models.}~~ElastoGen supports both explicit and implicit models. We dense-samples the implicit neural field, directly generating physically accurate dynamics without a simulator, enabling image-to-image generation from novel camera poses.}
  \label{fig:nerf}
\end{figure}
However, this approach requires many RNN loops to achieve effective global relaxation. This is because local operations are more effective in processing locally concentrated strains, while object-wise global deformation can only be progressively approximated by information exchange across voxels. This is also a well-known limitation in numerical computation --- Gauss-Seidel- or Jacobi-style iterative methods are less effective in relaxing low-frequency residual errors, which are often paired with a multigrid solver for large-scale problems. 

To address this issue, we apply SVD to encode Eq.~\eqref{eq:global} into a low-frequency latent space and directly solve the system within this space. The result is then decoded back to the original space, serving as the initialization for RNN-2. However, since the global matrix varies across objects and timesteps, performing frequent SVD computations on the global matrix is inefficient. Given that only a prediction of the overall low-frequency deformation is required, we approximate the object's global matrix using the global matrix of a uniform rasterized grid. This projection can be precomputed, eliminating the need for recalculation at each step or for each object. With this approach, each latent mode takes the form of a smooth sine wave, as illustrated in Fig.~\ref{fig:illustration}~(c).

\section{Experiments}
We implement ElastoGen using \texttt{Python}. Specifically, we use \texttt{PyTorch}~\citep{imambi2021pytorch} to implement the network and a simulator for training data generation. Our hardware is a desktop computer equipped with an \texttt{Intel} \texttt{i7-12700F} CPU and an \texttt{NVIDIA} \texttt{3090} GPU. Detailed statistics of the settings, models, and fitting errors are reported in Tab.~\ref{tab:statistics}. \emph{All the experiments are also available in the supplemental video}.

\begin{table*}[ht]
    \centering
    \small
    % \captionsetup{font=small}
    % \vspace{5pt}
    \begin{tabular}{cccccccccccc}
        \toprule
        \textbf{Scene} & \textbf{Grid resolution} & \textbf{$\#$DoFs} & \textbf{$\#$latent} & \textbf{$\Delta t$} & \textbf{$\#$R1} & \textbf{$\#$R2} & \textbf{EM} & \textbf{Fitting error} & \textbf{$t/\text{frame}$}\\
        \hline
        % \makecell{\textbf{ShapeNet}~(App.\ref{sec:more_exp})} & $32\times 32\times 32$ & $5\text{K}$ & $36$ & $0.002$ & $10$ & $213$ & NH & $1.32\times 10^{-4}$ & $0.08$\\
        % \hline
        \makecell{\textbf{Cantilever}~ (Fig.~\ref{fig:rods-bend})} & $16\times 3\times 3$ & $432$ & $18$ & $0.001$ & $5$ & $108$ & All & $4.11\times 10^{-4}$ & $0.01$\\
        \hline
        % \makecell{\textbf{Cantilever}~(Fig.~\ref{fig:rods-twist})} & $16\times 3\times 3$ & $432$ & $18$  & $0.001$ & $15$ & $140$ & NH & $9.67\times 10^{-5}$ & $0.01$\\
        % \hline
        \makecell{\textbf{Lego}~(Fig.~\ref{fig:nerf})} & $26\times 46\times 30$ & $11\text{K}$ & $54$ & $0.005$ & $15$ & $320$ & NH & $2.34\times 10^{-4}$ & $0.44$\\
        \hline
        \makecell{\textbf{Drums}~(Fig.~\ref{fig:nerf})} & $28\times 22\times 34$ & $4\text{K}$ & $54$ & $0.005$ & $15$ & $320$ & CR & $7.63\times 10^{-5}$ & $0.21$\\
        \hline
        \makecell{\textbf{Bridge}~(Fig.~\ref{fig:teaser})} & $66\times 13\times 27$ & $7\text{K}$ & $81$ & $0.003$ & $5$ & $96$ & StVK & $5.78\times 10^{-4}$ & $0.92$\\
        \hline
        \makecell{\textbf{Ship}~(Fig.~\ref{fig:teaser})} & $53\times 33\times 16$ & $14\text{K}$ & $81$ & $0.001$ & $5$ & $100$ & NH & $2.34\times 10^{-4}$ & $1.20$\\
        \hline
        \makecell{\textbf{monster}~(Fig.~\ref{fig:gen})} & $32\times 30\times 22$ & $20\text{K}$ & $36$ & $0.001$ & $5$ & $100$ & NH & $2.34\times 10^{-4}$ & $0.78$\\
        \hline
        \makecell{\textbf{shoe}~(Fig.~\ref{fig:gen})} & $48\times 30\times 20$ & $28\text{K}$ & $36$ & $0.001$ & $5$ & $100$ & NH & $2.34\times 10^{-4}$ & $1.08$\\
        % \hline
        \bottomrule
    \end{tabular}
    % \vspace{-6pt}
    \caption{\textbf{Experiments statistics.}~~We report detailed settings of our experiments. \textbf{$\#$DoFs}: the average number of DOFs involved in the optimization. $\Delta t$: the size of timestep. \textbf{$\#$R1}: the average loop count of RNN-1 for each step. \textbf{$\#$R2}: the average number of RNN-2 loops for each timestep. \textbf{\# latent}: the dimension of latent layer in the subspace encoder. \textbf{EM}: the elastic materials including Neo-Hookean (NH), StVK, and co-rotational (CR) models. \textbf{Fitting error}: the loss of NeuralMTL in Eq.~\eqref{eq:loss}. \textbf{$t/\text{frame}$}: the seconds needed for each frame. } 
         
    \label{tab:statistics}
\end{table*}

% \subsection{4D generation for any shapes}

% ElastoGen generates 4D elastic dynamics of 3D models with any shapes. To demonstrate this, we conduct experiments on multiple models from ShapeNet~\citep{chang2015shapenet} with arbitrary external forces and boundary conditions. Some results of ElastoGen are shown in Fig.~\ref{fig:shapenet}, and more are available in the appendix. All 3D objects are rasterized with a $32\times32\times32$ grid, which also serve as our subspace encoding. Cabinets are fixed at the bottom, twisted, and then released to yield elastic oscillations. Towers and plants sway under prescribed wind fields. Airplanes are pinned at the middle. Users apply sharp dragging force at the tip of the wings, resulting in interesting and realistic dynamic effects. These results show that different boundary conditions and external forces produce plausible dynamic outcomes. 
% \begin{figure*}
%   \centering 
%   \includegraphics[width =0.96\linewidth]{fig/demo_shapenet/shapenet.pdf}
%   \caption{\textbf{ElastoGen on ShapeNet.}~~ElastoGen generates physically grounded 4D dynamics for objects of any geometries. To demonstrate this property, we run ElastoGen for a wide range of 3D objects in ShapeNet with different boundary conditions and external forces. This figure shows snapshots of a subset of our results including cabinets (green), towers (blue), plants (yellow), and airplanes (red). These experiments are under the rasterization resolution of $32\times32\times32$.}
%   \label{fig:shapenet}
% \end{figure*}

\subsection{Quantitative validation of NeuralMTL}
\begin{figure*}[ht]
  \centering
  \begin{tabular}{c@{}c@{}c}
    \includegraphics[width=0.33\linewidth]{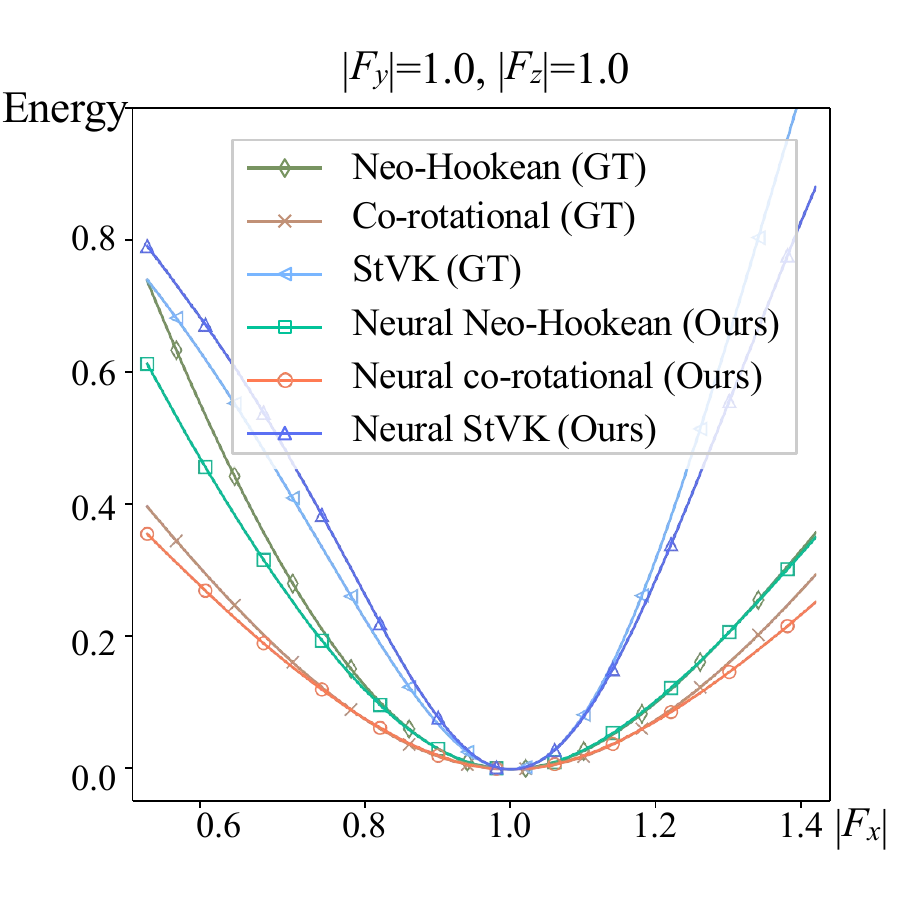} & 
    \raisebox{10pt}{\includegraphics[width=0.29\linewidth]{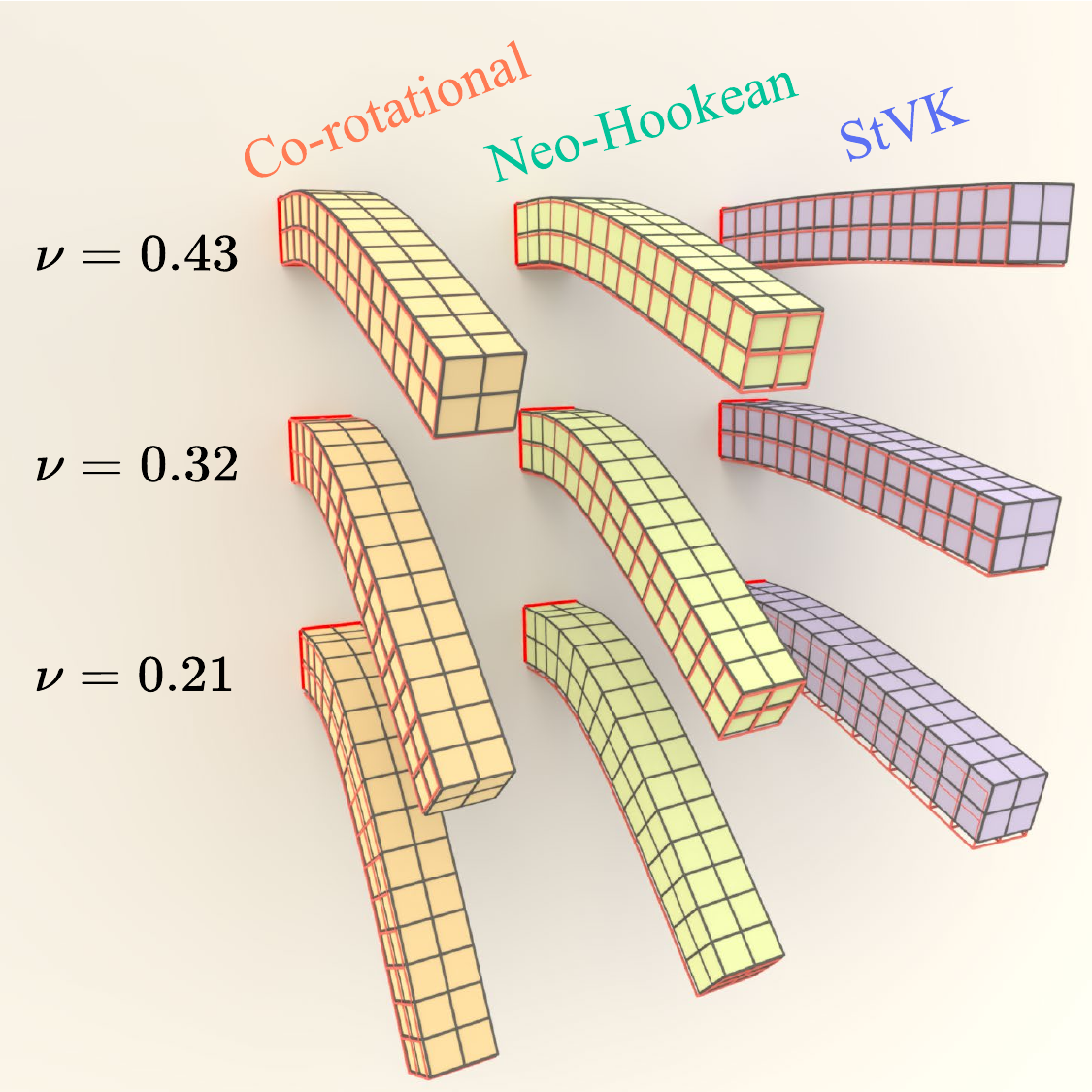}} &
    \includegraphics[width=0.33\linewidth]{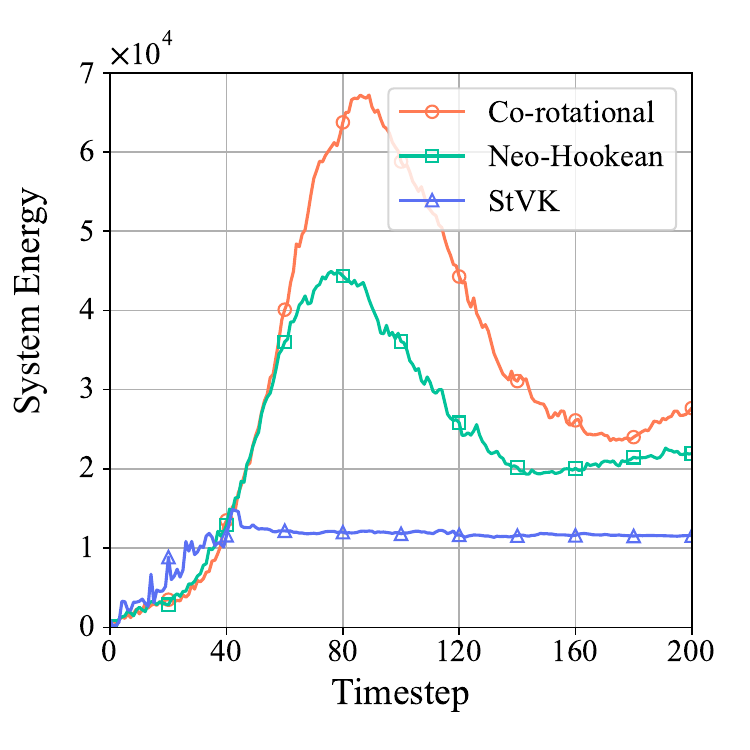}
    \\
    \small{\textbf{(a)} } &
    \small{\textbf{(b)} } & 
    \small{\textbf{(c)} }\\
  \end{tabular}
  \caption{\textbf{Quantitative validation of NeuralMTL.}~~\textbf{(a)}~Comparison between the energy computed from NeuralMTL strain and the ground truth energy. \textbf{(b)}~Visualization of the final static state of the cantilever beam generated by our method, with different materials and material parameters. \textbf{(c)}~Plots of the elastic energy during the prediction.}
  \label{fig:rods-bend}
\end{figure*}

ElastoGen replicates the behavior of complex hyperelastic materials with varying parameters. We quantitatively compare ElastoGen's results with those from the finite element method (FEM) using a cantilever beam bending test. ElastoGen predicts the trajectories for three classic materials co-rotational~\citep{brogan1986corotation}, Neo-Hookean~\citep{wu2001nh}, and StVK~\citep{barbivc2005stvk}. Each material is tested at three Poisson's ratios, with a fixed Young's modulus. (Poisson's ratio alters the material response more nonlinearly than Young's modulus). The results of ElastoGen, as shown in Fig.~\ref{fig:rods-bend} (b), align well with the results obtained from the classic method of FEM. Both overlap nearly perfectly. Such superior accuracy is due to our NeuralMTL prediction. As shown in Fig.~\ref{fig:rods-bend} (a), the diffusion-generated strain from NeuralMTL closely matches the ground truth (GT) with the correlation coefficient $r$ being larger than $0.98$ (calculated as $r=\frac{\sum_{i=1}^n\left(g_i-\bar{g}\right)\left(f_i-\bar{f}\right)}{\sqrt{\sum_{i=1}^n\left(g_i-\bar{g}\right)^2 \sum_{i=1}^n\left(f_i-\bar{f}\right)^2}}$ for each sample point $f_i$ and $g_i$ on neural strain and the ground truth curve, and $\bar{f}$ and $\bar{g}$ are their averages). We also plot the total neural energy variation over time for those materials ($\nu = 0.32$) in Fig.~\ref{fig:rods-bend} (c).

\begin{figure}[ht]
\centering
    \includegraphics[width = \linewidth]{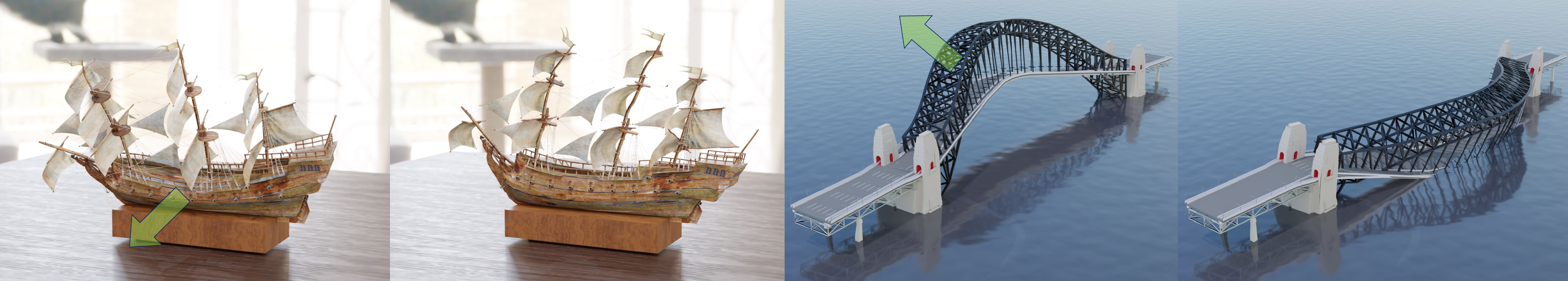}
    % \captionsetup{type=figure}
    \caption{\textbf{Experiments on high-resolution mesh objects.}~~ElastoGen generates realistic nonlinear elastic dynamics under external forces. Both high-frequency local details and low-frequency global deformations are effectively captured with a carefully designed nested RNN architecture.}
    \label{fig:teaser}
\end{figure}

\subsection{Versatility}

ElastoGen is a general-purpose generative AI model. As long as a 3D object can be rasterized, ElastoGen deals with both explicit, \eg, as shown in Fig.~\ref{fig:teaser}, and implicit shape representations. For instance, when ElastoGen readily takes an implicit neural radiance field (NeRF)~\citep{mildenhall2021nerf} based model. One can conveniently employ the Poisson-disk sampling as described in~\citet{feng2023pie} to obtain the rasterized model. Given user-specified external forces or position constraints, ElastoGen generates its further dynamics directly via a neural network without resorting to an underlying physic simulator as used in PIE-NeRF~\citep{feng2023pie}. Similarly, a 3DGS (3D Gaussian splatting)-based model~\citep{kerbl20233d} can also feed to ElastoGen for 4D generation. We demonstrate this using NeRF dataset images in Fig.~\ref{fig:nerf}.

ElastoGen enables artists and animators to quickly produce high-quality 4D animations, even for complex models. Examples of high-resolution, triangle mesh objects are shown in Fig.~\ref{fig:teaser}, where ElastoGen generates visually accurate dynamics while preserving fine structural details. Additionally, material parameters can be inversely learned from video to ensure consistency with observed dynamics.

\begin{figure}
    \centering

   \begin{minipage}{\linewidth}
        \centering
        \begin{minipage}{\linewidth}
            \centering
            \begin{minipage}{0.19\linewidth}   %left column
                \centering
                \includegraphics[width=\linewidth]{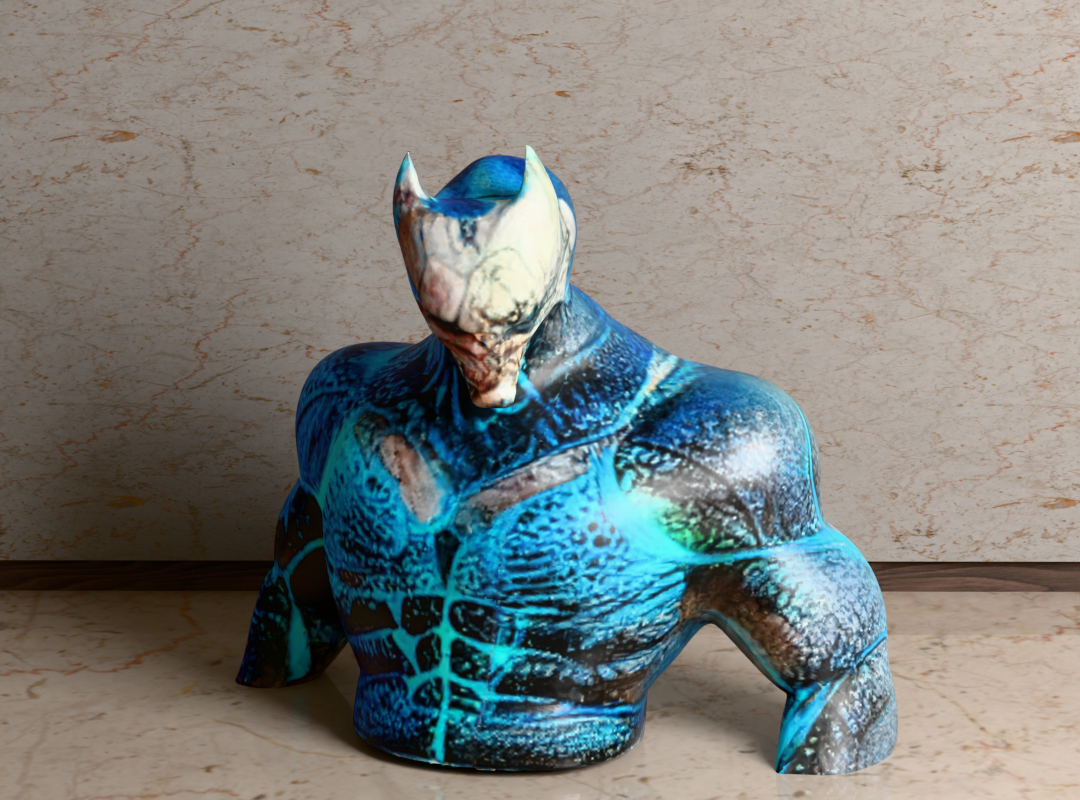}
                \includegraphics[width=\linewidth]{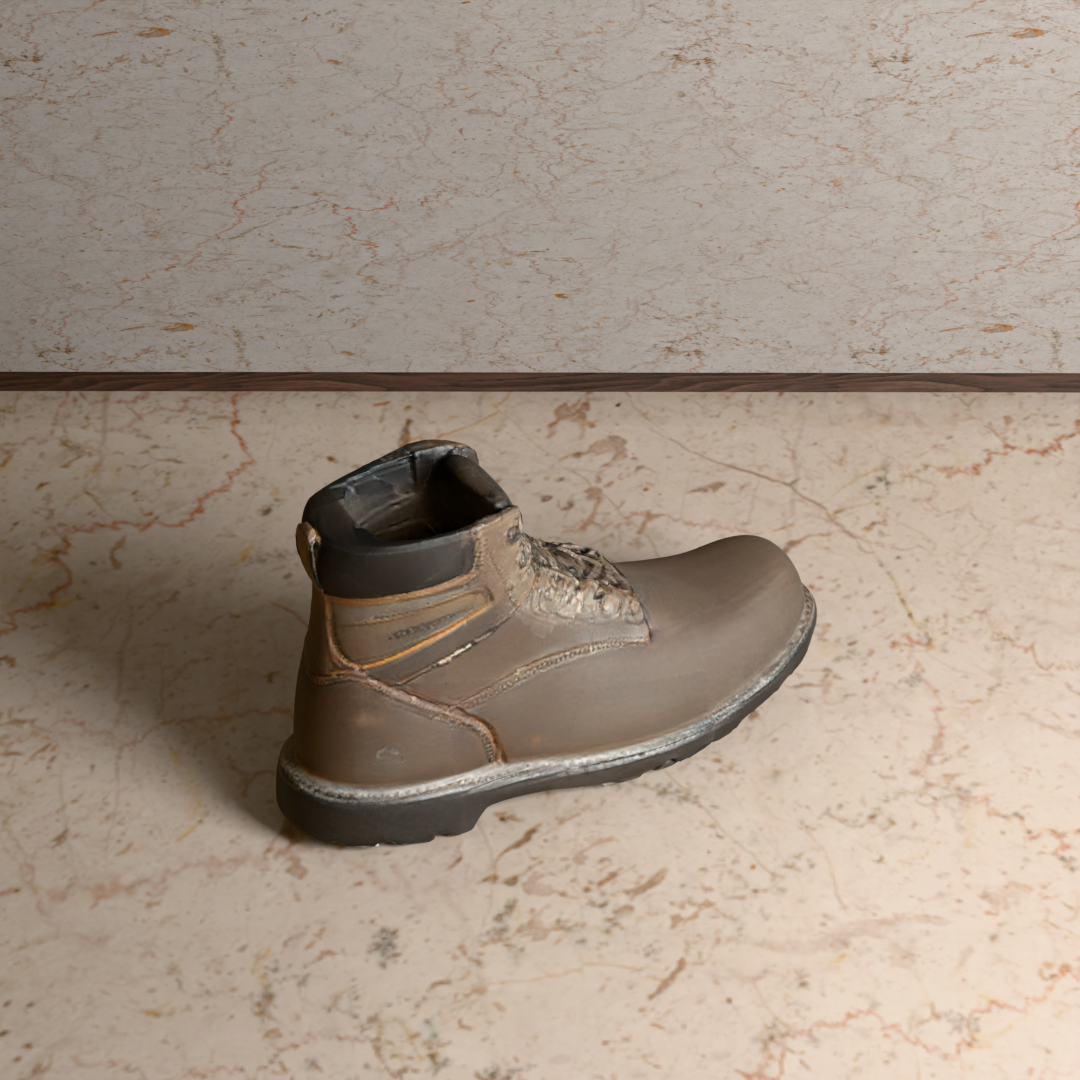}
                
            \end{minipage}
            \begin{minipage}{0.19\linewidth} %right column
                \centering
                \includegraphics[width=\linewidth]{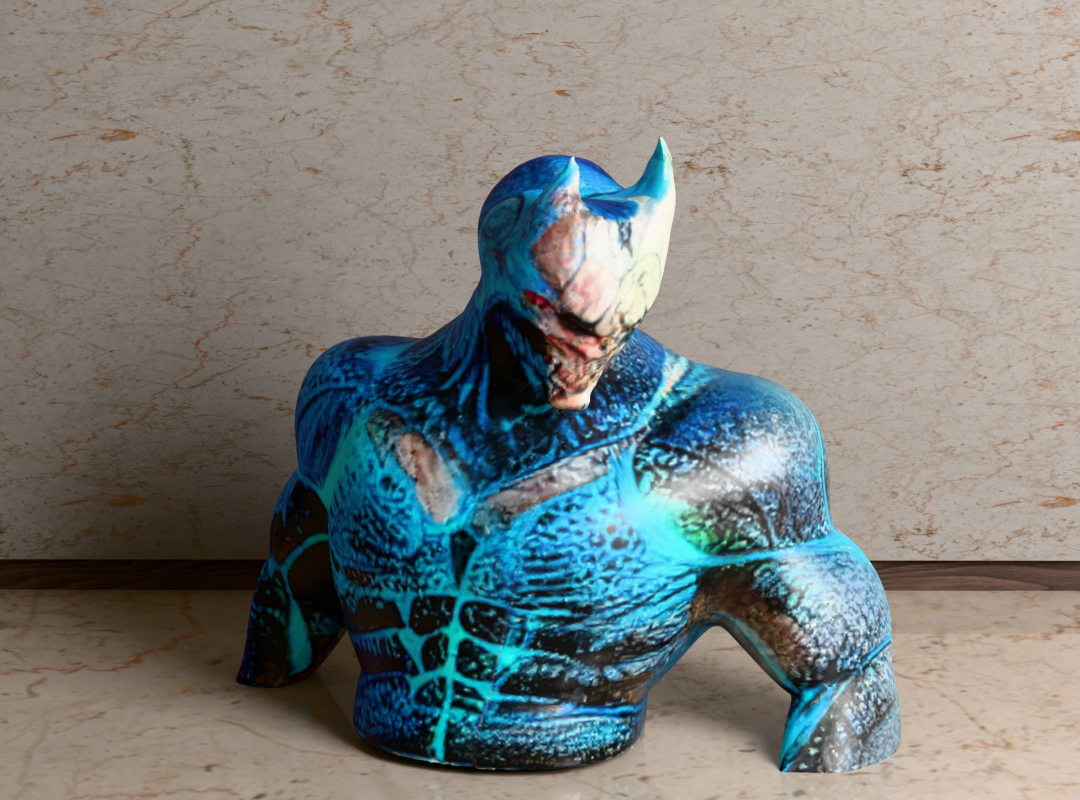}
                \includegraphics[width=\linewidth]{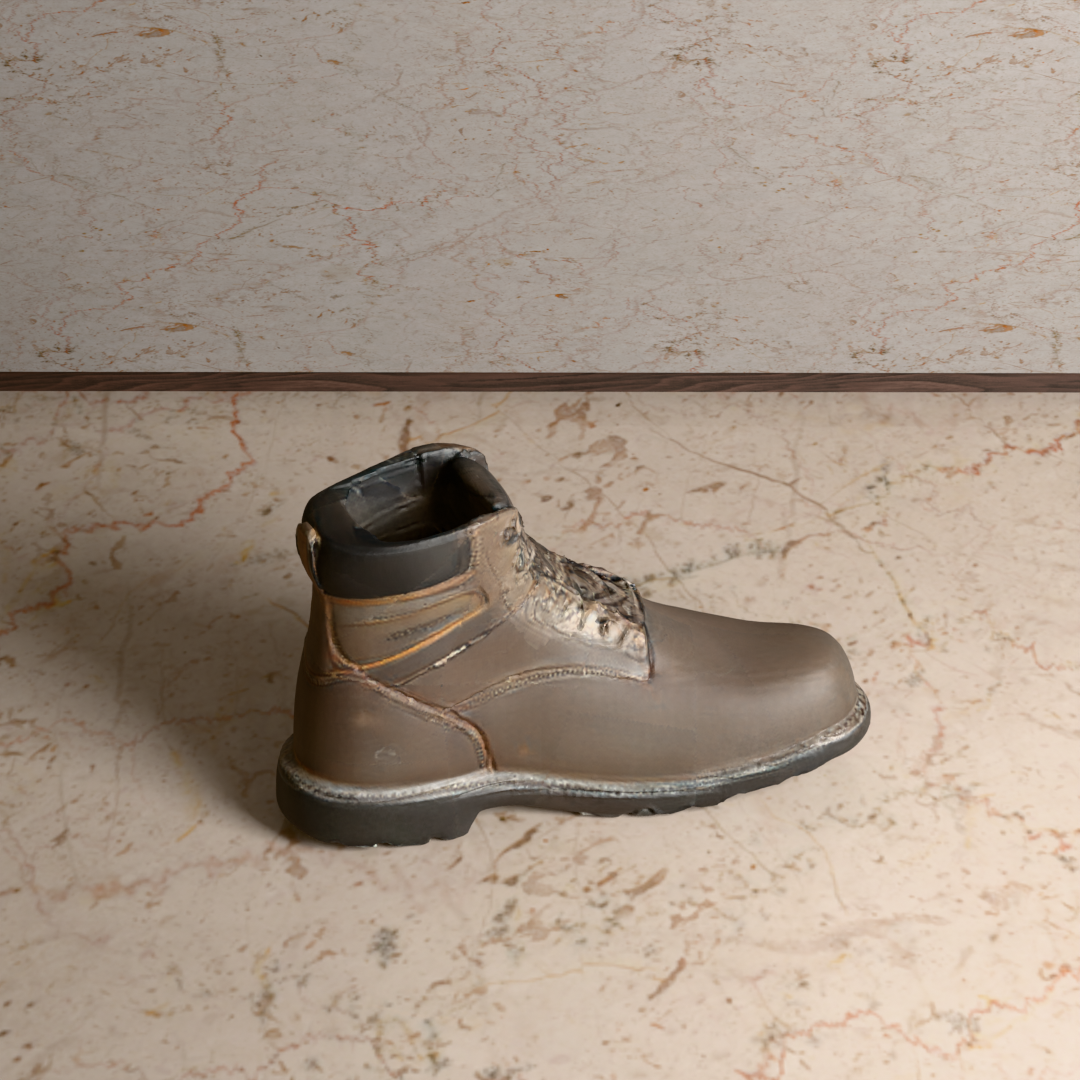}
            \end{minipage}
            \begin{minipage}{0.19\linewidth} %right column
                \centering
                \includegraphics[width=\linewidth]{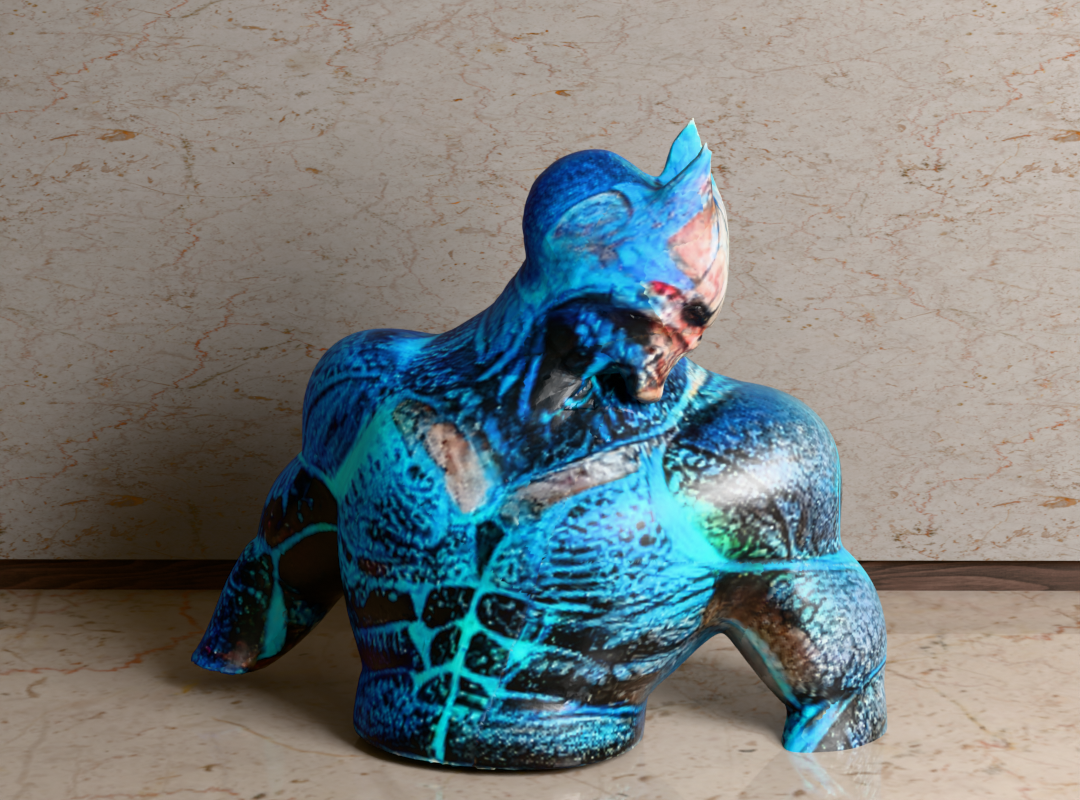}
                \includegraphics[width=\linewidth]{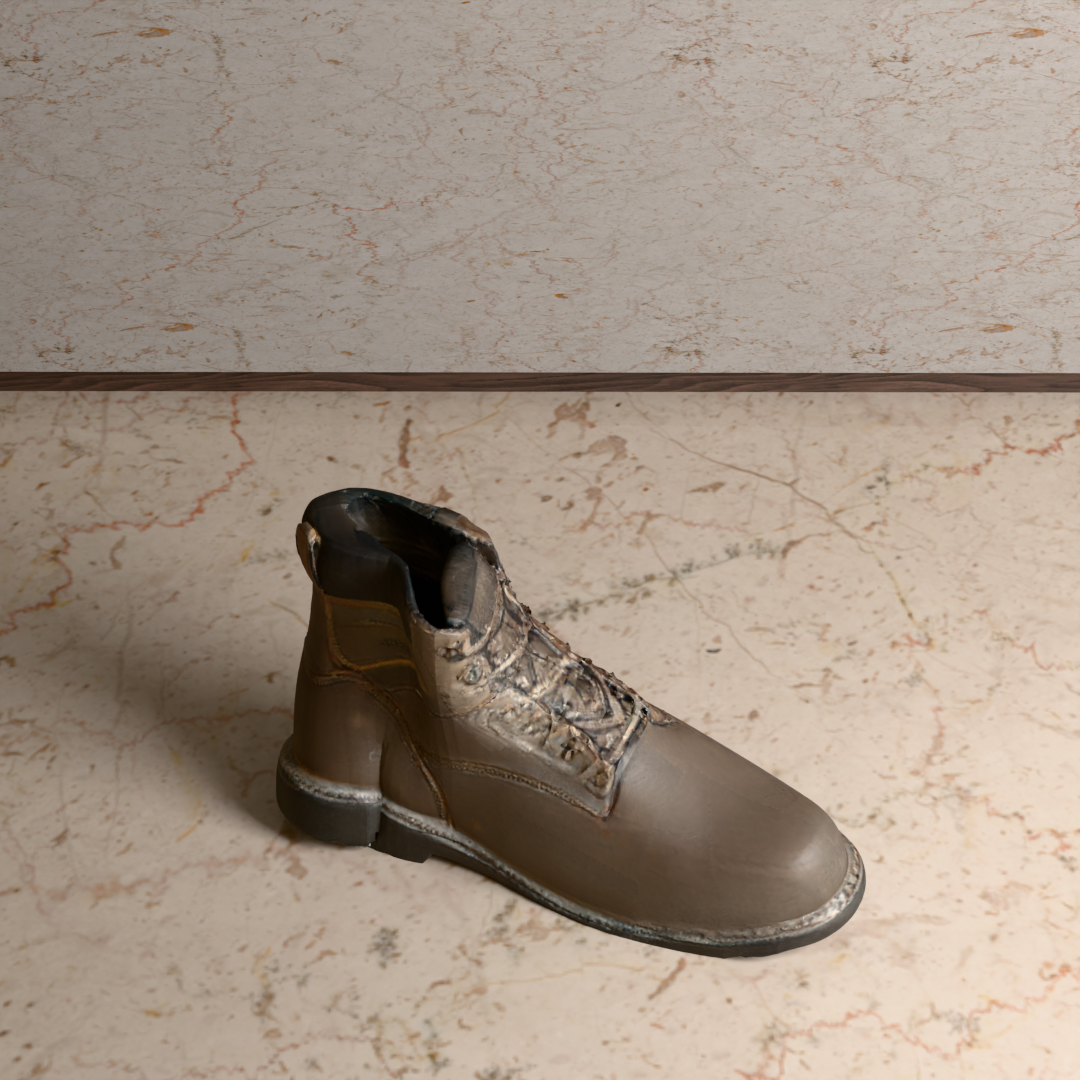}
            \end{minipage}
            \begin{minipage}{0.19\linewidth} %right column
                \centering
                \includegraphics[width=\linewidth]{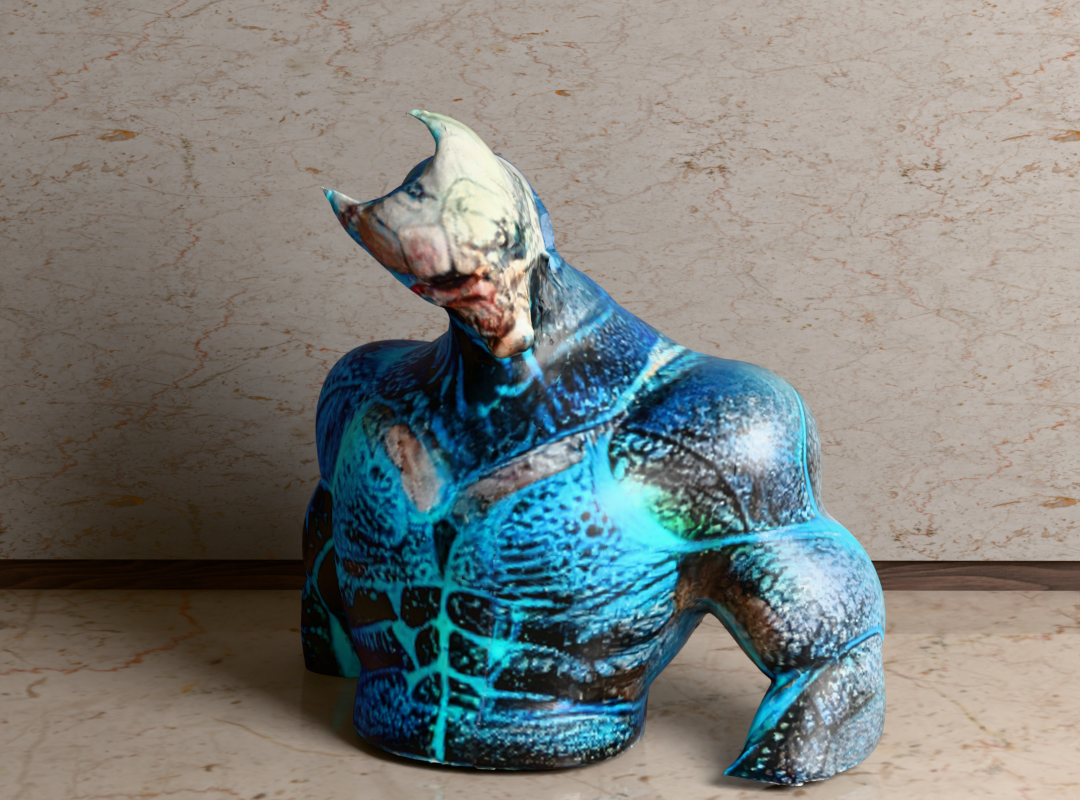}
                \includegraphics[width=\linewidth]{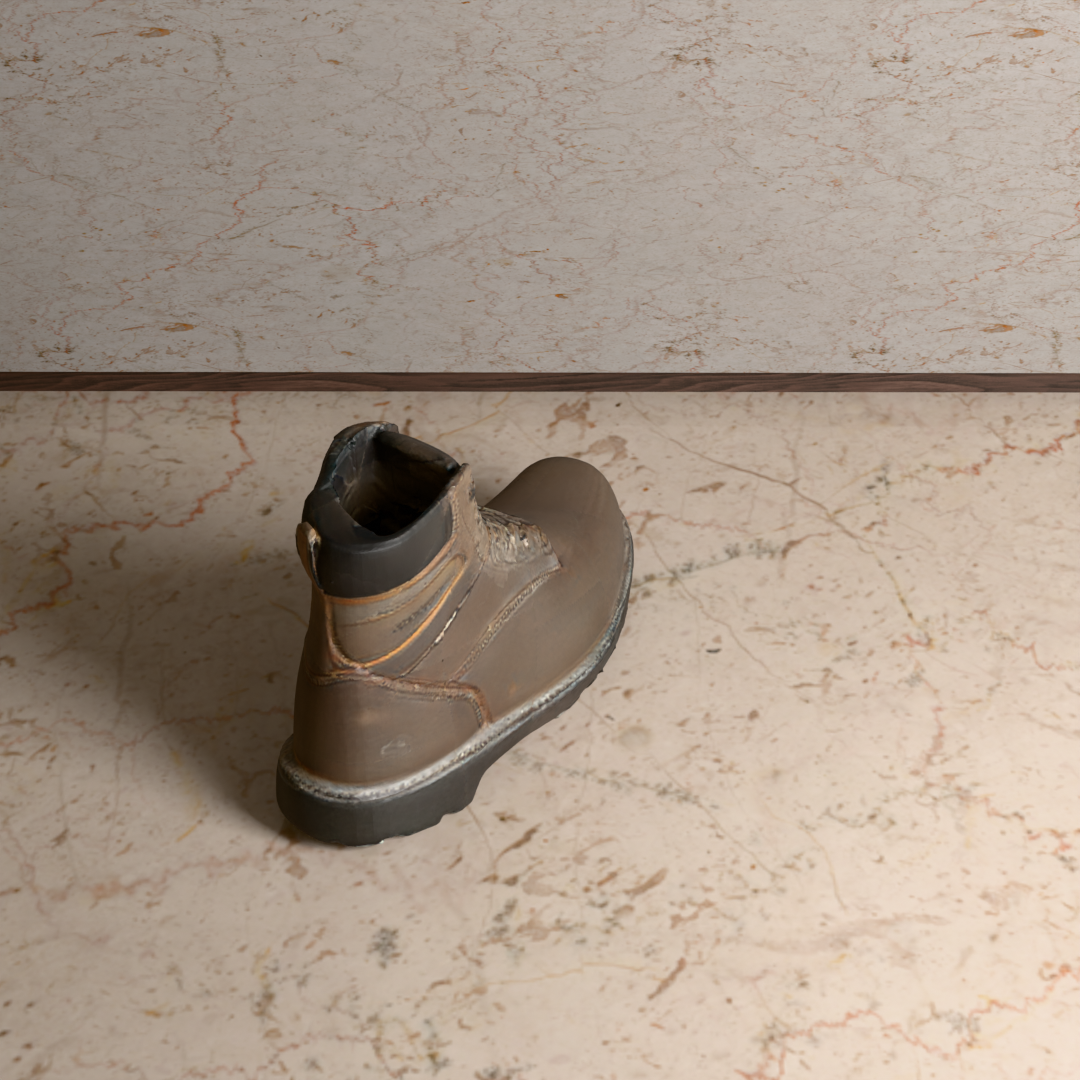}
            \end{minipage}
            \begin{minipage}{0.19\linewidth} %right column
                \centering
                \includegraphics[width=\linewidth]{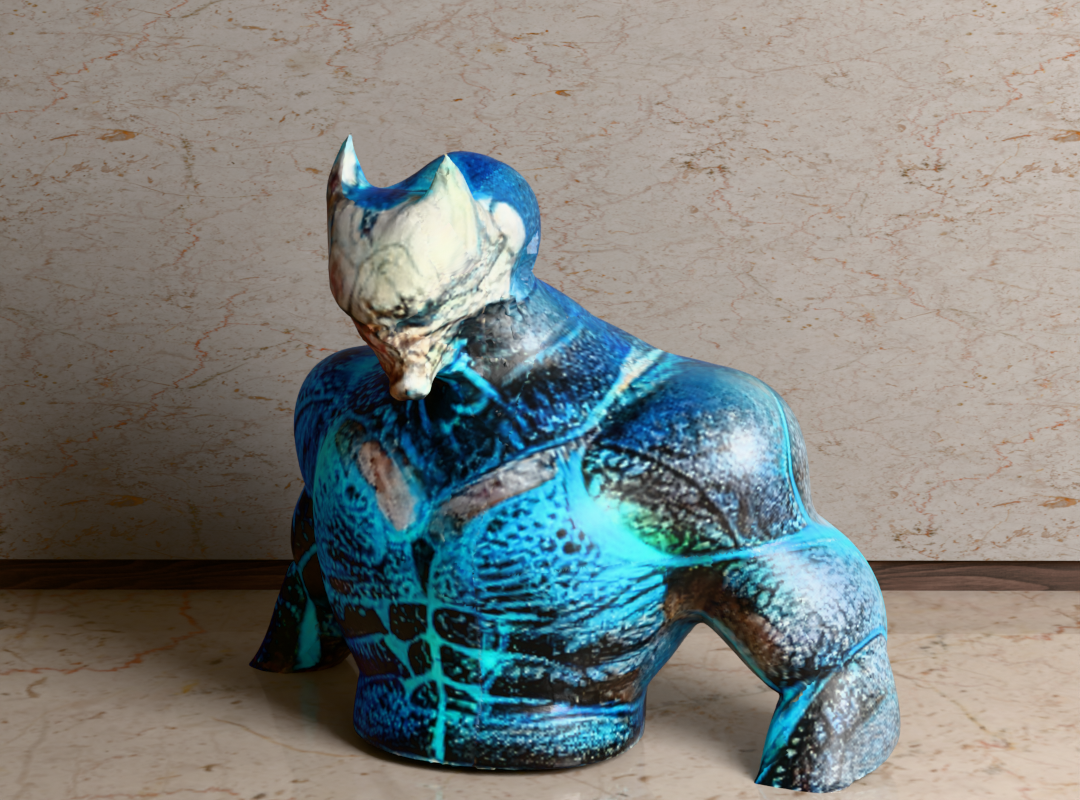}
                \includegraphics[width=\linewidth]{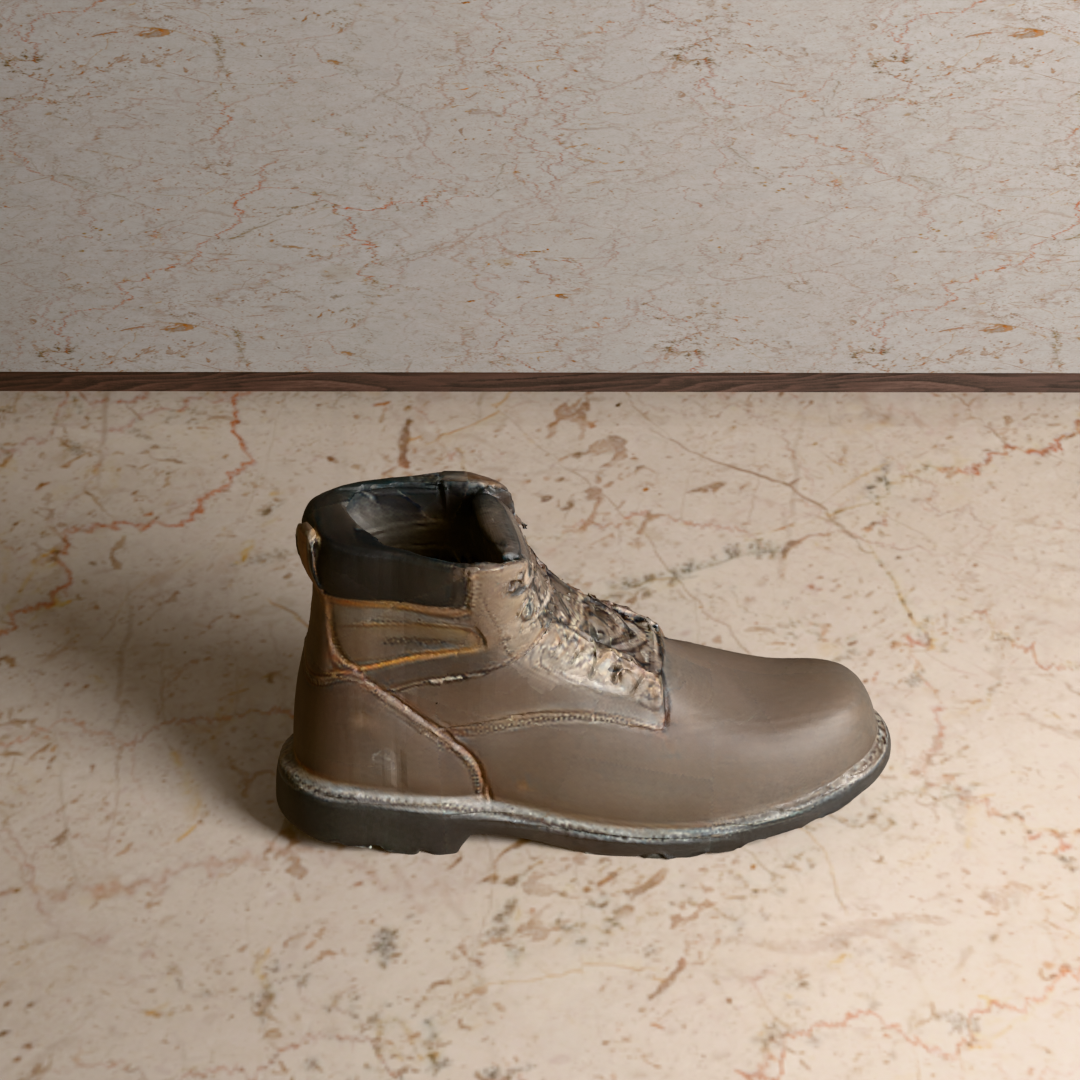}
            \end{minipage}
        \end{minipage}
    \end{minipage}

    \caption{\textbf{4D generation with ARM~\citep{feng2024arm}.}~~We combine ElastoGen with a 3D generation model ARM~\citep{feng2024arm} to perform 4D elastodynamics generation. The first row shows a monster shaking its head, while the second row shows a shoe being squeezed. }
    \label{fig:gen}
\end{figure}

ElastoGen can also serve as a downstream model for 3D generation frameworks to accomplish true 4D generation tasks. After rasterizing the 3D object generated by the upstream model, it can be used as the input to ElastoGen, which then produces physically plausible animations driven by forces or constraints. As shown in Fig.~\ref{fig:gen}, by integrating ElastoGen with ARM~\citep{feng2024arm}, we achieve high-quality and physically accurate 4D elastodynamics generation.

%he video to make the generation consistent with the observation.

\subsection{More comparisons}
\noindent \textbf{Comparison with ground truth.}
In addition to Fig.~\ref{fig:rods-bend}, we further compare ElastoGen with the FEM simulation under large-scale nonlinear twisting. The comparison is based on the Neo-Hookean material. For highly nonlinear instances, the physical accuracy of ElastoGen relies on the RNN loops --- more loops at both RNN-1 and RNN-2 effectively converge ElastoGen to the ground truth. Nevertheless, for general-purpose generation, fewer iterations also yield good results. Detailed experimental results and error plots are provided in the supplementary materials.

\begin{figure}[ht]
\includegraphics[width = \linewidth]{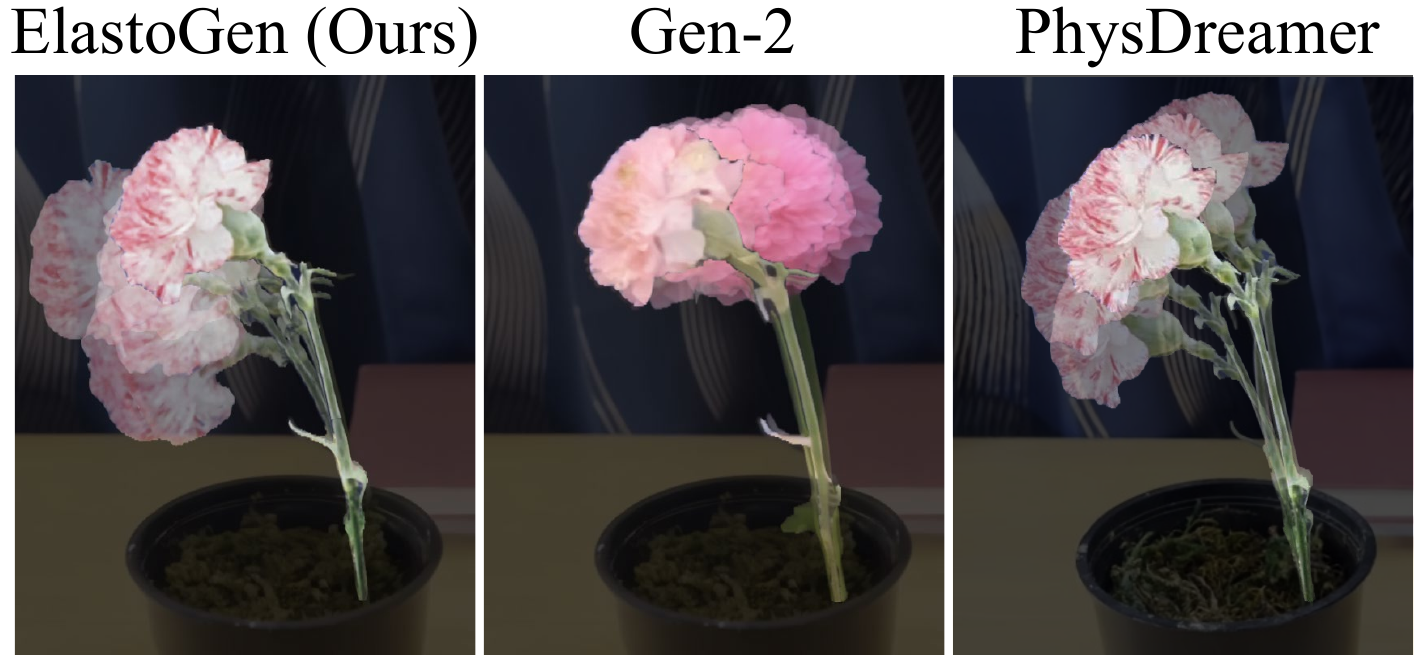}
\caption{\small \textbf{Comparison (trajectory) between ElastoGen, Gen-2~\citep{runwayml2024image2video_gen2} and PhysDreamer~\citep{zhang2024physdreamer}.}~~We visualize the trajectory of a swinging carnation using ElastoGen, Gen-2, and PhysDreamer. Note that PhysDreamer can only produce plausible elastodynamics with tiny time steps.}
\label{fig:traj_comp}
\end{figure}

\noindent \textbf{Comparison with SOTA competitors.}~We compare ElastoGen with existing 4D generative models, including Gen-2~\citep{runwayml2024image2video_gen2} and PhysDreamer~\citep{zhang2024physdreamer} in Fig.~\ref{fig:traj_comp}, showing that ElastoGen excels in both physical accuracy and geometric consistency. Gen-2 generates moderate motion with limited nonlinearity, such as rotation and bending, and fails to maintain geometric consistency over time, 
causing changes in both color and shape. 
This is a common issue for observation-based 4D models, where complex visual correlations in training data are difficult to decouple in a monolithic deep model. PhysDreamer produces plausible elastodynamics only at small time steps ($\Delta t<6.0\times10^{-5})$ due to the underlying explicit integration, which is known to be unstable under large time steps. 
In contrary, ElastoGen is able to generalize on large time steps. In Tab.~\ref{tab:quant_comp}, we present a quantitative comparison of error using the Intersection over Union (IoU) metric between ElastoGen, Gen-2~\citep{runwayml2024image2video_gen2}, and PhysDreamer~\citep{zhang2024physdreamer}. The reference data is generated using \cite{feng2023pie}. Our method demonstrates superior accuracy in comparison to the others.

\begin{table}
  
    \centering
    \small
    \begin{minipage}[t]{0.98\linewidth}
        \centering
        \begin{tabular}{ccc}
            \toprule
            \quad\quad\textbf{ElastoGen (Ours)}\quad\quad & \textbf{Gen-2} & \quad\quad\textbf{PhysDreamer}\quad\quad\\
            \midrule
            94\% & 64\% & 75\% \\
            \bottomrule
        \end{tabular}
    \end{minipage}
    \captionof{table}{\small \textbf{Comparison of quantitative error between ElastoGen, Gen-2 and PhysDreamer.}~~We compute the Intersection over Union (IoU) using reference data generated by \cite{feng2023pie}. Higher IoU values indicate greater accuracy.}
    \label{tab:quant_comp}
\end{table} 

% The results and convergence plots are shown in Fig.~\ref{fig:rods-twist}. In this standard test, one end of the beam is fixed, and ElastoGen predicts its twisting trajectory under external forces. We note that 50 RNN loops converge ElastoGen prediction to GT. Aggressively decreasing the loop count to 20 still yields satisfactory results. In contrast, 1, 3, and 5 iterations result in noticeably stiffer dynamics. In this experiment, RNN-2 uses an 18-dimension subspace encoder to extract low-frequency residuals. Without the encoding, local relaxation fails to converge.  

\section{Conclusion}

ElastoGen is a knowledge-driven deep model that embeds physical principles and numerical procedures into the network design. As a result, it is remarkably lightweight and compact. Each module is designed for a specific computational task aimed at minimizing the total variational energy. This modular design enables decoupled training, removing the need for large-scale training datasets. The accuracy of ElastoGen can be easily controlled by NeuralMTL, which predicts the current strain based on observed numerical computations.

ElastoGen also has limitations. It currently lacks collision support and is less efficient for thin geometries due to excessive convolution operations on empty voxels. It may also fail to converge with highly stiff materials, such as near-rigid objects. Future improvements will include adding dynamics for more physical phenomena, integrating collision support, and automating the setting of physical parameters to enable real-world dynamics with minimal input.

{\small
\bibliography{aaai2026}
}

\clearpage
\renewcommand*{\thesection}{\Alph{section}}
\setcounter{section}{0}
\setcounter{page}{1}
\appendix

\section{Supplemental video}
We refer the readers to the supplementary video to view the animated results for all examples.

\section{Diffusion network}

The goal is to train a diffusion network $\mathcal{D}$ to generate the weights $\mathbf{W}$ of a corresponding NeuralMTL $\mathcal{N}$, given the material parameters $\{e, \nu\}$. Here, $\mathbf{W}$ denotes the weights of $\mathcal{N}$, and the process is formulated as a conditional diffusion problem guided by $\{e, \nu\}$, such that $\mathbf{W} = \mathcal{D}(e, \nu)$.

To this end, we first construct a dataset consisting of 1000 paired samples of $\{e, \nu\}$ and $\mathbf{W}$, as described in \S~\ref{sec:diffusion}. Following the approach of Wang \etal~\cite{wang2024neural}, we utilize latent diffusion models (LDM)~\cite{rombach2022high} to generate $\mathbf{W}$, as our preliminary experiments show that directly learning $\mathbf{W}$ leads to suboptimal performance. To address this, we train an autoencoder to map the network weights $\mathbf{W}$ to a 256-dimensional latent vector, in which the diffusion process is performed.

When training the diffusion model, the autoencoder remains fixed, serving solely to encode $\mathbf{W}$ into its latent representation $l$. At each diffusion timestep $t$, we introduce noise $\epsilon_t$ to $l$, resulting in $l_t = l + \epsilon_t$. The objective is to train a noise prediction model, $\epsilon_\theta(l_t, t; e, \nu)$, to estimate the noise $\epsilon_t$ at each timestep $t$, as described in \S~\ref{sec:bg_diffusion}. During inference, we begin with random noise and progressively remove noise from it using the noise prediction model $\epsilon_\theta$, guided by the material parameters $\{e, \nu\}$. This iterative denoising process produces a 256-dimensional latent vector, which is subsequently passed through the decoder to generate the corresponding network weights $\mathbf{W}$. 

We train the autoencoder using a learning rate of $1 \times 10^{-3}$ and the diffusion model with a learning rate of $1 \times 10^{-4}$. Both models are trained for 1000 epochs with a batch size of 64. The architecture of the autoencoder and diffusion model is detailed in Tab.~\ref{tab:diffusion_network_details}. Note that in diffusion process the 256-dimensional latent vector is viewed as a 1-channel $16\times16$ image.

\begin{table*}[h!]
\centering
% \scriptsize
\begin{tabularx}{0.85\linewidth}{X c c c} % X is for a column that stretches
\toprule
\textbf{Network} & \textbf{Layers} & \textbf{$\#$Output features} & \textbf{Description} \\ 
\midrule
\multirow{2}{*}{Autoencoder}       & FC   & 8192, 4096, 2048, 1024, 512, 256  & Encoder \\ 
      &  FC  & 512, 1024, 2048, 4096, 8192, 17153  & Decoder    \\ 
\midrule
\multirow{4}{*}{Diffusion model}      & Conv2D      & 256, 512 & Down-sample     \\ 
& FC & 256 & Time embedding \\
& FC & 256 & $\{e,\nu\}$ embedding \\
& Conv2D & 256, 1 & Up-sample \\
\bottomrule
\end{tabularx}
\caption{\textbf{Architecture of the autoencoder and diffusion model}. FC denotes the fully connected layer, and Conv2D represents the 2D convolution layer. The third column refers to the number of output features in each layer.}
\label{tab:diffusion_network_details}
\end{table*}

\section{Convolutional deformation gradient}\label{subsec:deformation_gradient}
Given an input 3D object, ElastoGen rasterizes it into a set of 3D voxels. For $i$-th voxel, ElastoGen uses a 3D CNN to calculate $\mathcal{G}_i$. As $\mathcal{G}_i$ has an analytic format as described in Eq.~\eqref{eq:analytic_operator}, the kernel's weights of 3D CNN can be directly computed. In details, for $i$-th voxel containing 8 vertices, let $\mathbf{A}_i = [\mathbf{q}_1, \mathbf{q}_2,...\mathbf{q}_8]\in\mathbb{R}^{3 \times 8}$ and $\bar{\mathbf{A}}_i = [\bar{\mathbf{q}}_1, \bar{\mathbf{q}}_2,...\bar{\mathbf{q}}_8]\in\mathbb{R}^{3 \times 8}$ be deformed and rest-shape position of the vertices respectively, the weights of 3D CNN can be filled with $\left[\big(\bar{\mathbf{A}} \bar{\mathbf{A}}^\top\big)^{-1}\bar{\mathbf{A}}\otimes\mathbf{I}\right]\in\mathbb{R}^{9 \times 24}$. Here, the 3D CNN has an input channel of $3$, an output channel of $9$ and a kernel size of $2\times2\times2$.

\section{Global phase}\label{subsec:global}
As stated in the main text, we need to solve the global linear system as in Eq.~\eqref{eq:global}, which requires determining $\mathbf{L}_i$ and $\mathbf{b}_i$. We abbreviate the neural strain $\mathcal{N}(\mathcal{G}_i[\mathbf{q}_i])$ as $\mathcal{N}$, rewriting Eq.~\eqref{eq:isotropy}, the energy $E_i$ for voxel $i$ is
\begin{equation}\label{eq:Ei}
    E_i = \frac{\omega_i}{2}\left\|\mathbf{F}_i\mathcal{N}-\mathbf{U}_i\mathbf{V}_i^\top\right\|^2_{F}.
\end{equation}
For the convenience of subsequent derivations, we rewrite Eq.~\eqref{eq:Ei} as:
\begin{equation}\label{eq:Ei_vec}
\begin{split}
    E_i = &\frac{\omega_i}{2}\left\|\mathbf{N}\mathrm{vec}(\mathbf{F}_i)-\mathrm{vec}(\mathbf{R}_i)\right\|^2_{F}\\
    =&\frac{\omega_i}{2}\left\|\mathbf{N}\mathbf{G}_i\mathbf{q}-\mathrm{vec}(\mathbf{R}_i)\right\|^2_{F},\\
\end{split}
\end{equation}
where $\mathbf{N}=\mathcal{N}^\top\otimes \mathbf{I}$, $\mathrm{vec}(\cdot)$ flattens a matrix into a vector, $\mathbf{G}_i$ is a linear operator projects DOFs $\mathbf{q}$ to the i-th element's deformation gradient $\mathbf{F}_i$, and $\mathbf{R}_i=\mathbf{U}_i\mathbf{V}_i^\top$. Taking the derivative of Eq.~\eqref{eq:Ei_vec} with respect to position $\mathbf{q}$ we obtain
\begin{equation}\label{eq:rhs_global}
    \frac{\partial E_i}{\partial \mathbf{q}}=\omega_i\left (\mathbf{G}_i^\top\mathbf{N}^\top\mathbf{N}\mathbf{G}_i\mathbf{q} - \mathbf{G}_i^\top\mathbf{N}^\top\mathrm{vec}(\mathbf{R}_i)\right).
\end{equation}
Comparing it to the definition, $\mathbf{L}_i\mathbf{q}-\mathbf{b}_i:=\frac{\partial E_i}{\partial \mathbf{q}}$, we obtain the expression for $\mathbf{b}_i$ and $\mathbf{L}_i$ as
% We decompose the formula into a term linear in $\mathbf{q}$ and a term independent of $\mathbf{q}$ as follows:
\begin{equation}\label{eq:bili}
\mathbf{L}_i=\omega_i\mathbf{G}_i^\top\mathbf{N}^\top\mathbf{N}\mathbf{G}_i,\quad \mathbf{b}_i=\omega_i\mathbf{G}_i^\top\mathbf{N}^\top\mathrm{vec}(\mathbf{R}_i).
\end{equation}
As it indicates, for each voxel, we can obtain $\mathbf{b}_i$ by applying the transformation $\mathbf{G}_i^\top$ to $\mathbf{N}^\top\mathrm{vec}(\mathbf{R}_i)$. For $\mathbf{G}_i$ has been trained as a convolutional kernel as described in \S~\ref{subsec:deformation_gradient}, we can directly fetch the previously trained kernel and perform transposed 3D convolution. 

For the linear system in Eq.~\eqref{eq:global}, we rewrite it as $\mathbf{A}\mathbf{q}=\mathbf{b}$ for brevity. For any diagonally dominant matrix $\mathbf{A}$, the linear system $\mathbf{A}\mathbf{q}=\mathbf{b}$ can be solved using iterative method as:
\begin{equation}\label{eq:jacobi}
    \mathbf{q}^{k+1}=\mathbf{D}^{-1}(\mathbf{b}-\mathbf{B}\mathbf{q}^k),
\end{equation}
where $\mathbf{D}$ is the diagonal part of $\mathbf{A}$ and the off-diagonal part $\mathbf{B}=\mathbf{A}-\mathbf{D}$, and $\mathbf{q}^k$ is the result after $k$ loops of RNN-2. In our case, $\mathbf{A}=\frac{\mathbf{M}}{h^2}+\sum_i \mathbf{L}_i$ and $\mathbf{b}=\mathbf{f}_q+\frac{\mathbf{M}}{h^2}\left(\mathbf{q}^n+h \dot{\mathbf{q}}^n\right)+\sum_i \mathbf{b}_i$ according to Eq.~\eqref{eq:global}. Note that we use superscript $n$ to indicate timestep and superscript $k$ as the index for RNN-2 loops.

\begin{figure*}[!ht]
% \vspace{-10pt}
  \begin{minipage}{0.98\linewidth}
    \raisebox{0pt}{\includegraphics[width=\linewidth]{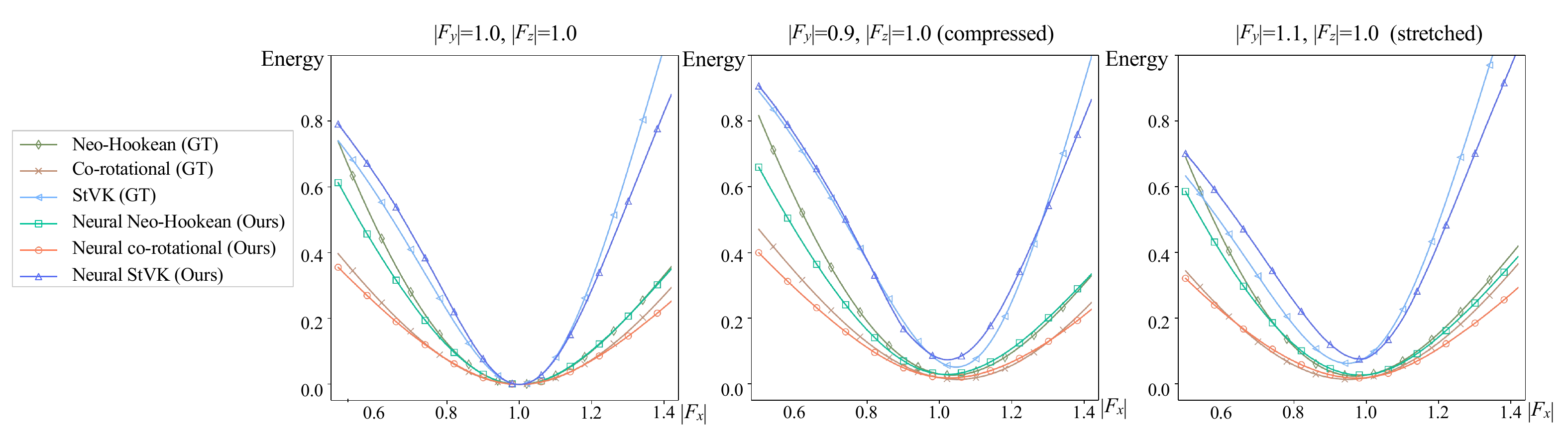}}
  \caption{\textbf{More quantitative validation of NeuralMTL.}~Comparison between the energy computed from NerualMTL strain and the ground truth under different configurations.}
  \label{fig:rods-more}
  \end{minipage}
\end{figure*}

Similar to \S~\ref{subsec:deformation_gradient}, to compute $\sum_i \mathbf{L}_i \mathbf{q}^k$, we first apply $\mathbf{G}_i$ to $\mathbf{q}^k$, then right-multiply the result with $\omega_i\mathbf{N}^\top\mathbf{N}$, and finally apply $\mathbf{G}_i^\top$ for each voxel. This process can be implemented using a 3D contolution, a matrix multiplication, and a transposed 3D convolution. The iterative process is formulated as a recurrent network (\ie, RNN-2) to solve the global system.

\section{Broader impact}
Our model integrates computational physics knowledge into the network structure design, significantly reducing the data requirements and making both the training and network structure more lightweight. It blends the boundaries of machine learning, graphics, and computational physics, providing new perspectives for network design. Our model does not necessarily bring about any significant ethical considerations.

% we know $\mathbf{A}$, and we understand that in the voxelized grid, only adjacent points have non-zero entries in $\mathbf{A}$. Therefore, the original Jacobi (Eq.~\eqref{eq:jacobi}) essentially uses $1$-ring neighbors' information. In Eq.~\eqref{eq:2ring_jacobi}, we are effectively using $2$-ring neighbors' information. Consequently, we can train a CNN to collect $2$-ring information for a point by using the $1$-ring neighbors' $1$-ring information. This CNN does not need real physical dynamics data for training; we can use completely random 1-ring neighbors information to train the CNN to collect the required information.

% For $\mathbf{L}_i$, we train a CNN to map $\mathbf{G}_i\mathbf{N}_i$ to $\mathbf{G_i}\mathbf{N}_i\mathbf{N_i}^\top\mathbf{G_i}^\top$, i.e., mapping any matrix $\mathbf{M}$ to $\mathbf{M}\mathbf{M}^\top$. We then incorporate the object's mass to get the global matrix. 

\section{More quantitative validations}
We compare the NeuralMTL strain with the ground truth under various deformed configurations as in Fig.~\ref{fig:rods-more}. In each case, the neural energy models closely match the ground truth, demonstrating the effectiveness and expressiveness of our neural approximations for these nonlinear energy functions.

\begin{figure*}[ht]
\vspace{-10pt}
\centering
  \begin{minipage}{\linewidth}
      \hspace{15pt}
      \begin{minipage}{0.3\linewidth} 
        \includegraphics[width=\linewidth]{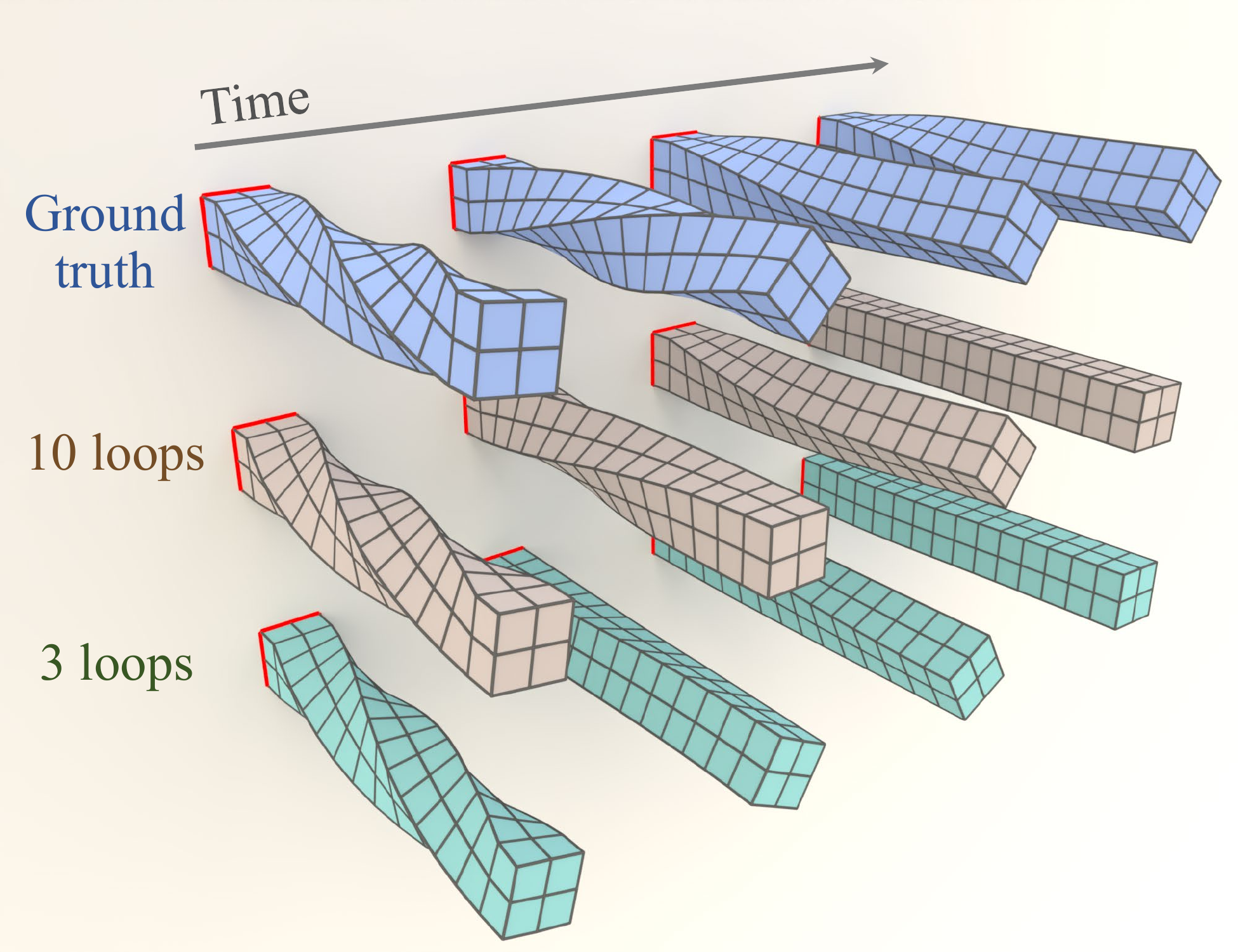}
        \put(-20,10){\scalebox{.9}{\color{black} (a)}}
        % \label{fig:rods-t}
      \end{minipage}
      \hspace{10pt}
      \begin{minipage}{0.6\linewidth}
        \includegraphics[width=\linewidth]{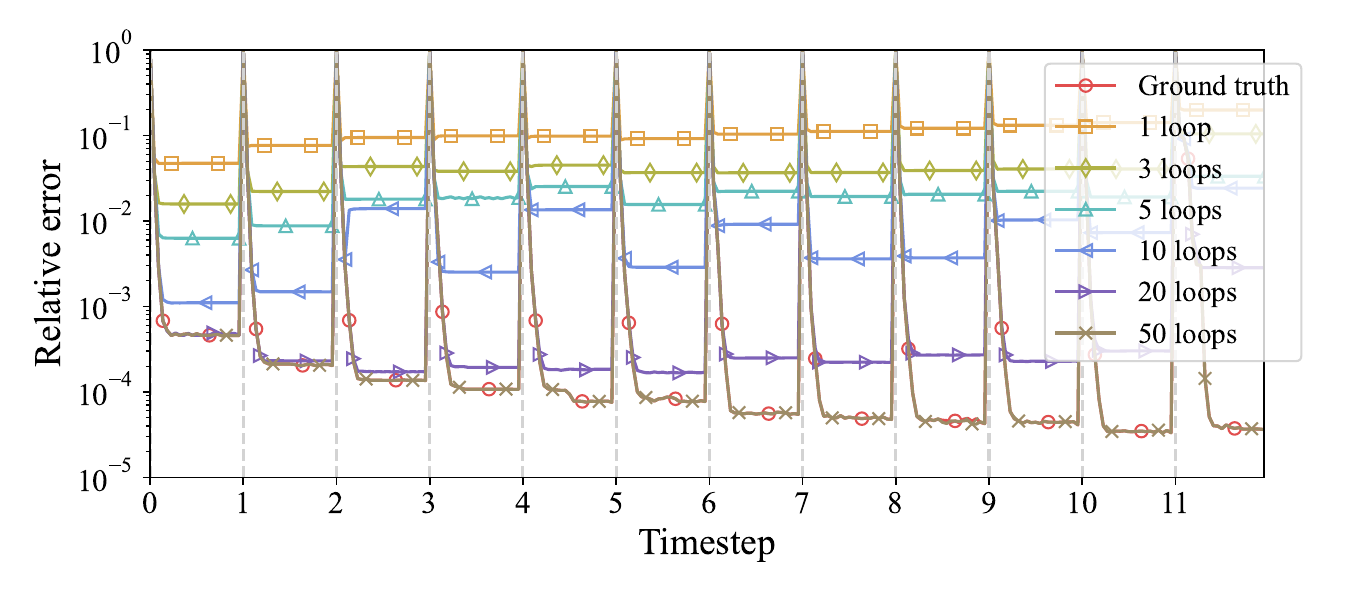}
        \put(-10,13){\scalebox{.9}{\color{black} (b)}}
        % \label{fig:error-time}
      \end{minipage}

      \caption{\textbf{Convergence for different RNN loops.}~~\textbf{(a)} Comparison with FEM with different RNN loops. We note that increasing RNN loops effectively converges ElastoGen to the ground truth. However, fewer loops also give good results in general. \textbf{(b)} Relative errors for under different RNN-1 loops for each timestep. An 18-dimension subspace encoder is used to extract low-frequency residuals.}
    \label{fig:rods-twist}
  \end{minipage}
\end{figure*}

\section{Convergence study}
To quantify the impact of RNN loops and the subspace encoding, we compare ElastoGen predictions using different RNN loops with the ground truth, computed via solving the global matrix with a direct solver, in terms of relative error. Specifically, the variational formulation requires that the derivative of the total energy (elastic + kinetic – external work) vanishes at equilibrium. Since the total derivative equals zero at convergence, we define the ground-truth residual as the negative of known external forces (\eg, gravity, boundary forces), and compare it to the predicted residual composed of the derivatives of elastic and kinetic energies.

The results and convergence plots are shown in Fig.~\ref{fig:rods-twist}. In this test, one end of the beam is fixed, and ElastoGen predicts its twisting trajectory under forces. We observe that 50 RNN iterations converge the ElastoGen prediction to GT. Reducing the loop count to 20 still yields satisfactory results, while using only 1, 3, or 5 iterations leads to noticeably stiffer dynamics. In this experiment, RNN-2 employs an 18-dimensional subspace encoder to extract low-frequency residuals. Without this encoding, local relaxation fails to converge.

\section{More experiments}\label{sec:more_exp}
\begin{figure*}[t]
  \centering 
  \includegraphics[width = \linewidth]{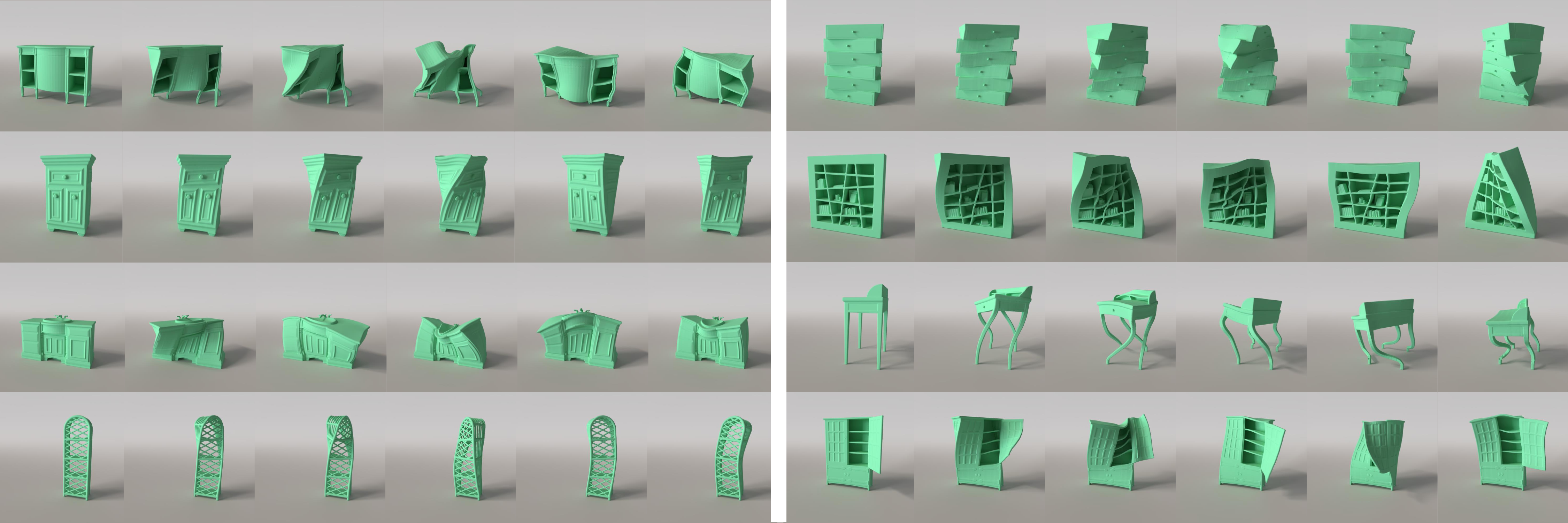}
  \includegraphics[width = \linewidth]{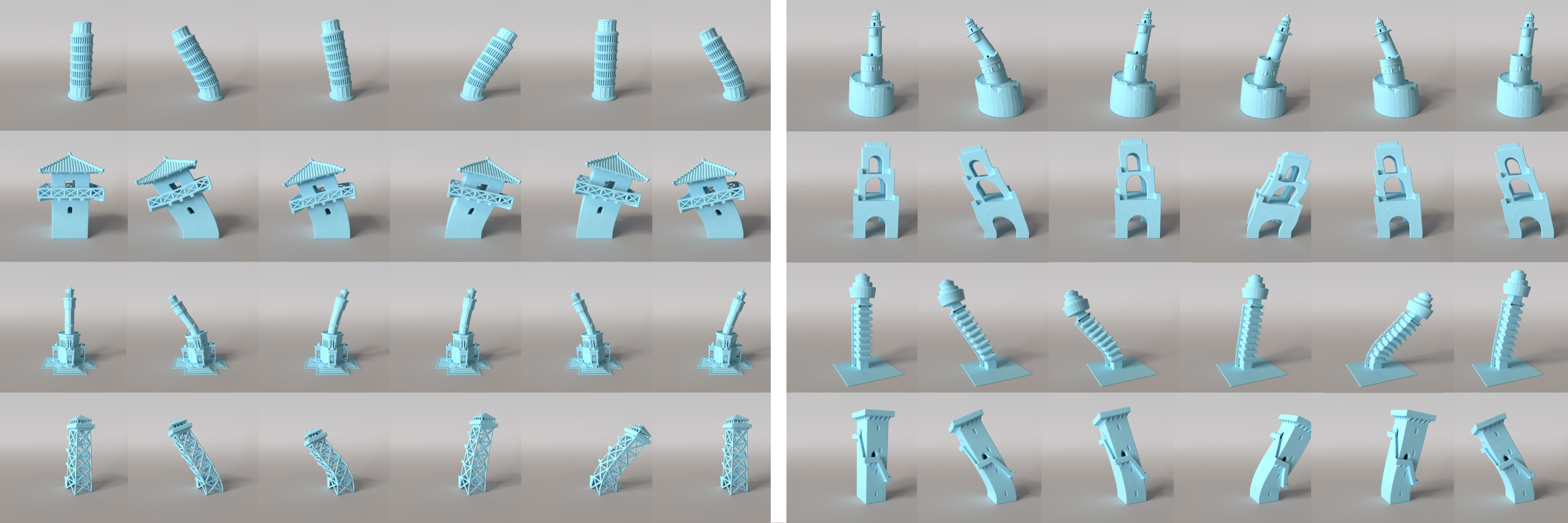}
  \includegraphics[width = \linewidth]{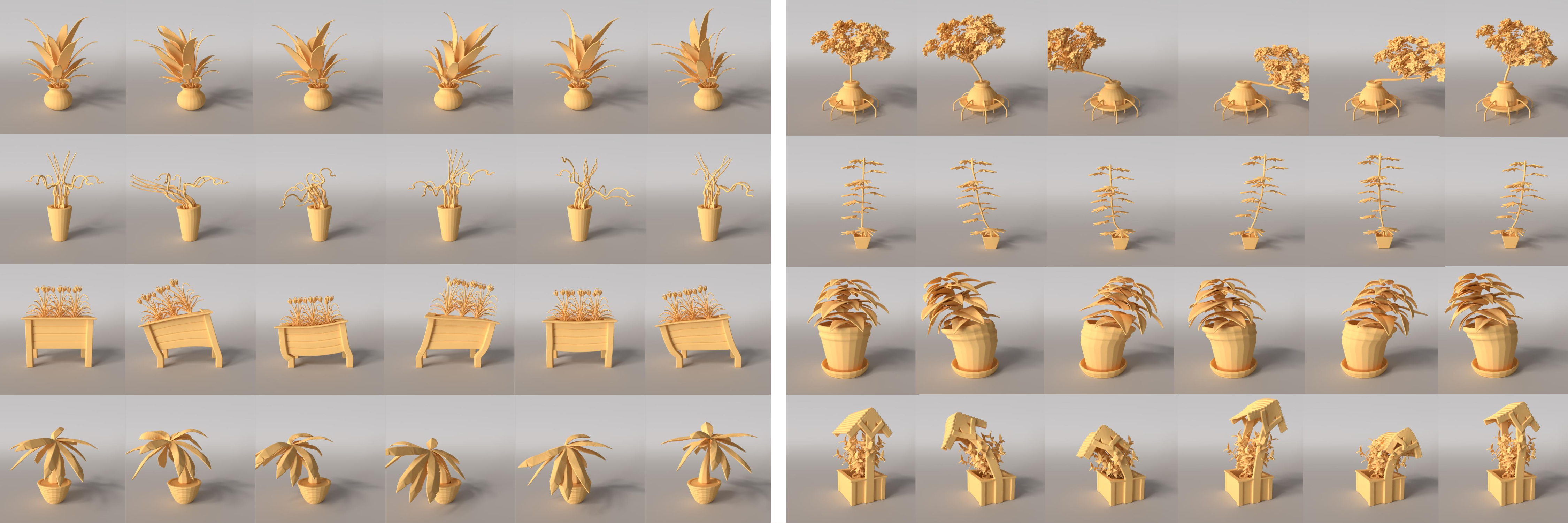}
  \caption{\textbf{Additional experiments on ShapeNet.}~Here are more results of cabinets, towers, and plants.}
  \label{fig:shapenet1}
\end{figure*}

\begin{figure*}[htbp]
  \centering 
  \includegraphics[width = \linewidth]{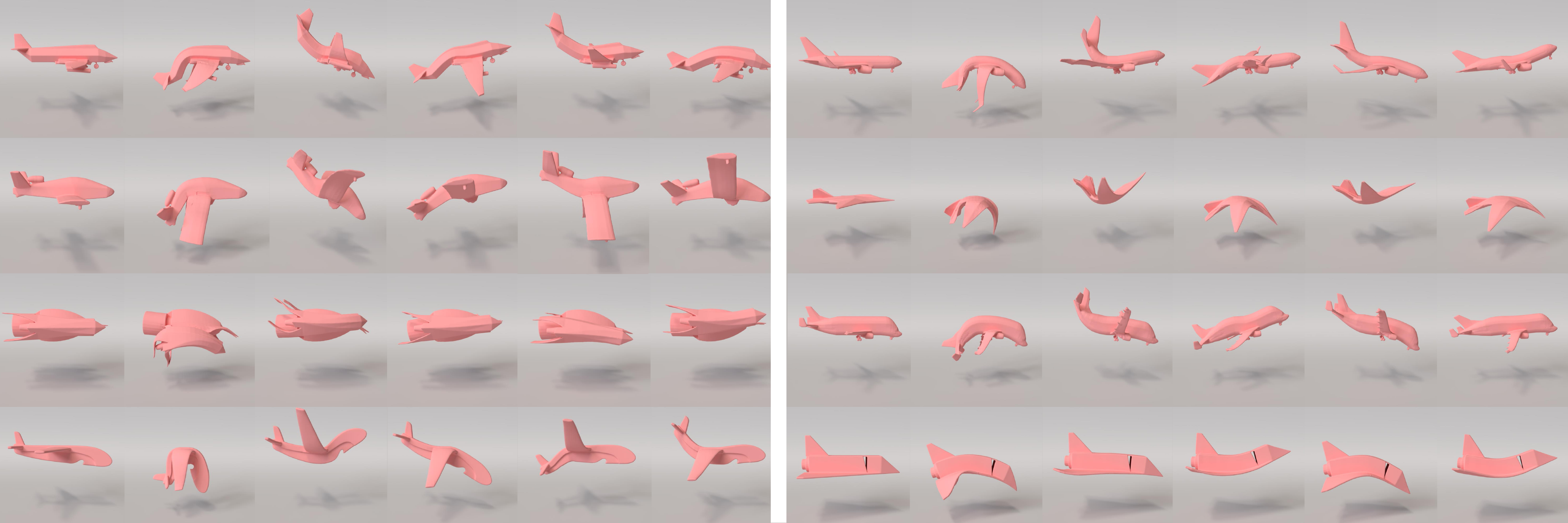}
  \includegraphics[width = \linewidth]{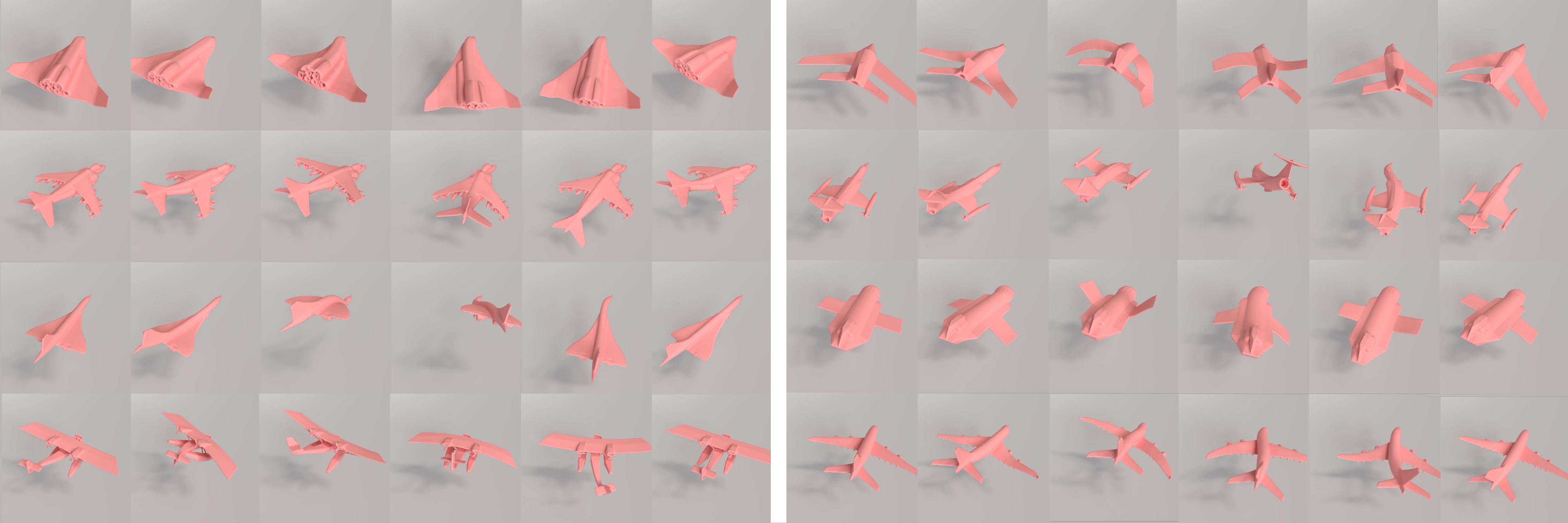}
  \includegraphics[width = \linewidth]{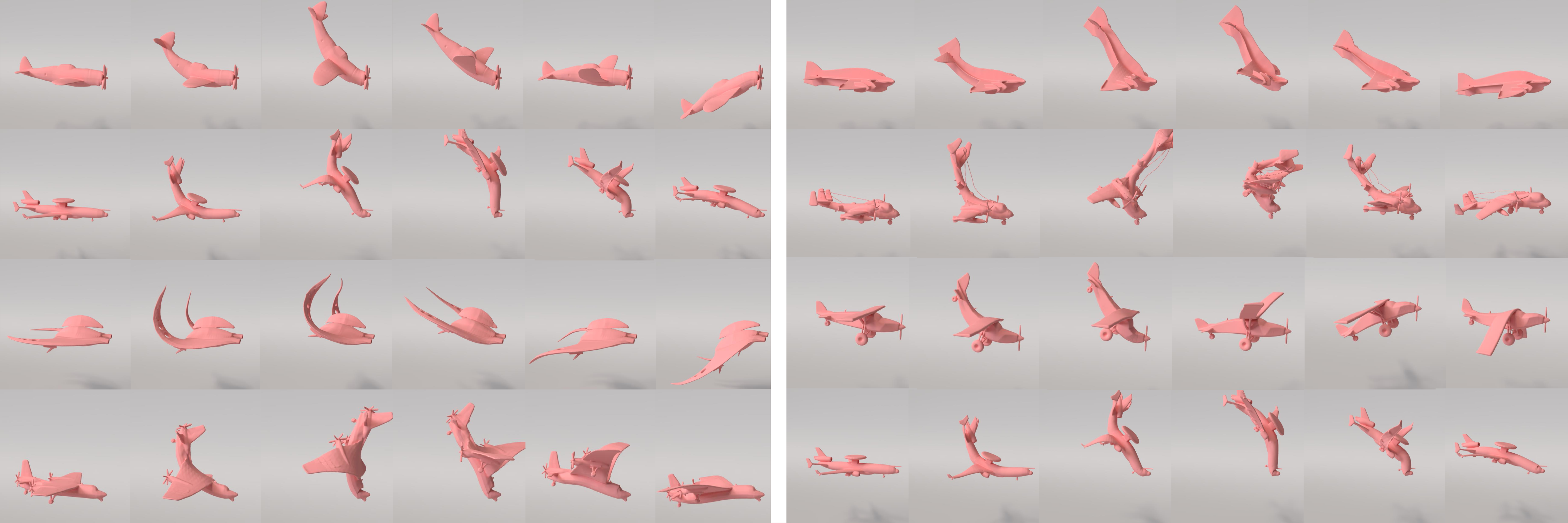}
  \caption{\textbf{Additional experiments on ShapeNet (continued).}~Here are more results of airplanes with different force and boundary settings.}
  \label{fig:shapenet2}
\end{figure*}

We provide additional results in Fig.~\ref{fig:shapenet1} and Fig.~\ref{fig:shapenet2} to demonstrate the robustness of ElastoGen. For animated results, we refer the readers to supplemental video.
\clearpage

\clearpage

\end{document}